\documentclass[12pt]{article}
\usepackage{aliases}
\graphicspath{{./figures/}}

\begin{document}

\title{Learning end-to-end inversion of circular Radon transforms in the partial radial setup}
\author{
Deep Ray{\footnote{(Corresponding Author)
deepray@umd.edu, Department of Mathematics,
University of Maryland, College Park, MD 20742, USA}} \and
Souvik Roy{\footnote{souvik.roy@uta.edu, Department of Mathematics, The University of Texas at Arlington, Arlington, TX 76019, USA}}
  }
  \date{}
\maketitle

\begin{abstract}
We present a deep learning-based computational algorithm for inversion of circular Radon transforms in the partial radial setup, arising in photoacoustic tomography. We first demonstrate that the truncated singular value decomposition-based method, which is the only traditional algorithm available to solve this problem, leads to severe artifacts which renders the reconstructed field as unusable. With the objective of overcoming this computational bottleneck, we train a ResBlock based U-Net to recover the inferred field that directly operates on the measured data. Numerical results with augmented Shepp-Logan phantoms, in the presence of noisy full and limited view data, demonstrate the superiority of the proposed algorithm.
    
\end{abstract}


\section{Introduction}
\begin{figure}[!htbp]
\centering
\includegraphics[width=0.5\textwidth ]{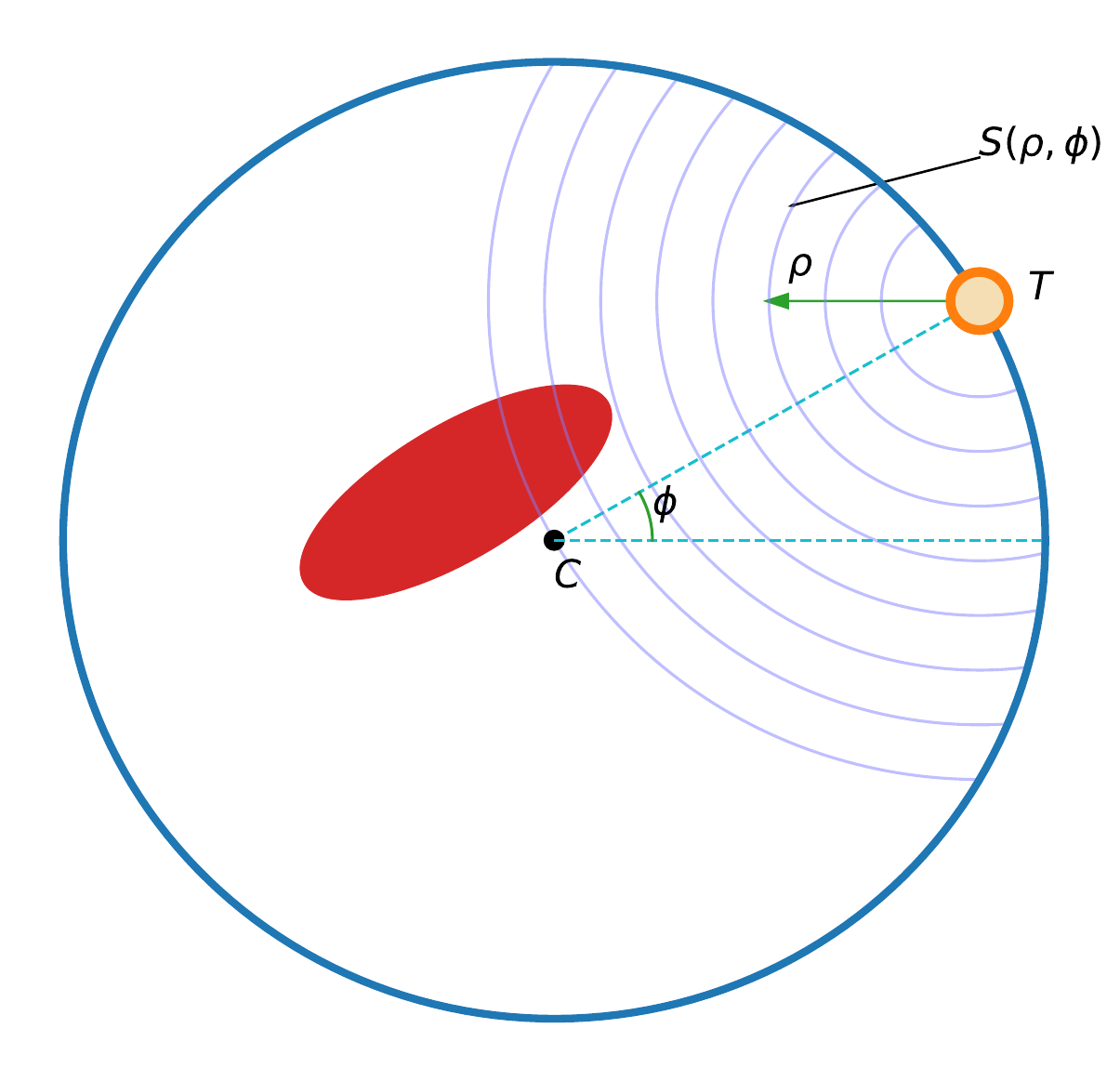}
\caption{Description of the PAT framework for a circular domain $\Omega$.}
\label{fig:setup}
\end{figure}

Photoacoustic tomography (PAT) is a hybrid imaging modality that uses a combination of light and sound waves to provide qualitative description of the properties inside a body (e.g, see \cite{Ammari10,Arridge2016,Elbau12,Nae14}). In PAT, a body is irradiated using a short pulse of electromagnetic (EM) laser light waves. The EM waves travel inside the body, represented by the domain $\Omega$ (see Figure \ref{fig:setup}), and a part of it is absorbed. This leads to a thermal expansion and contraction of the body, which generates acoustic wave pulses traveling throughout the body. Transducers $T$ placed on the boundary of the body $\partial\Omega$, also known as the acquisition domain, receive these acoustic wave signals. The measured signals are then used to determine the initial acoustic wave pressure distribution inside $\Omega$.

Since sound waves have very weak contrast inside tissues, under the assumption of a constant acoustic wave sound speed, the acoustic wave signals  measured by the transducers $T$ at any time can be considered as the
superposition of all the pulses reflected from the inhomogeneities, with the total distance traveled by the reflected
pulse being constant. Assuming that the two-dimensional cross-section of the body is a disk of radius $R$ with center $C$, the aforementioned superposition at a fixed time can be represented by a circular wavefront denoted by $S(\rho,\phi)$ having radius $\rho$ and with centre $(R\cos\phi, R\sin\phi)$ located on $\partial \Omega$. The acoustic wave signal data is collected over circular wavefronts of all possible radii $\rho \in (0,R)$ (corresponding to measurement times) and centres on $\partial \Omega$ at all possible polar angle locations $\phi \in [0,2\pi)$. This is known as the partial radial setup, in contrast to the traditional fully available radial data setup, where the data is available for $\rho \in (0,2R)$ (e.g., see \cite{Finch2004,Finch2007,Kunyansky2007}). The motivation of the partial radial setup is that, in practice, a partial radial data acquisition setup implies a low cost of obtaining the data in comparison to the fully radial data setup. Moreover, the partial radial data setup is different from the traditional partial or sparse data setup in PAT, in the sense that the traditional partial or sparse data setup involves using data that is measured on a portion of the acquisition domain (e.g., see \cite{Davoudi2019,Guan2019}). However, in this case, full data in the radial direction is considered. In contrast, our setup deals with data measured for a shorter time. Mathematically, in our setup, the initial pressure distribution must be recovered from partial measurement data along the radial direction.  

We denote the initial acoustic wave pressure distribution as a two-dimensional function $f(r,\theta)$ in polar coordinates, with $r\in[0,R]$ and $\theta \in [0,2\pi)$. Then, the data measured at the transducer $T$ is given by the circular Radon transforms, denoted by $g(\rho,\phi)$, and defined by \begin{equation}\label{eq:RadonTransform}
g(\rho,\phi) =  \int_{S(\rho,\phi)} f(r,\theta)~ds,
\end{equation}
where $ds$ is the parametrization of the arc length on the circle $S(\rho,\theta)$. The corresponding inverse problem in PAT can then be formulated as follows:
Given $g(\rho,\phi)$ for $\rho\in(0,R)$ and $\phi \in [0,2\pi)$, find $f(r,\theta)$ in $\Omega$ by solving the integral equation \eqref{eq:RadonTransform}.

\begin{figure}[H]
\centering
\subfloat[Full view]{\includegraphics[width=0.4\textwidth ]{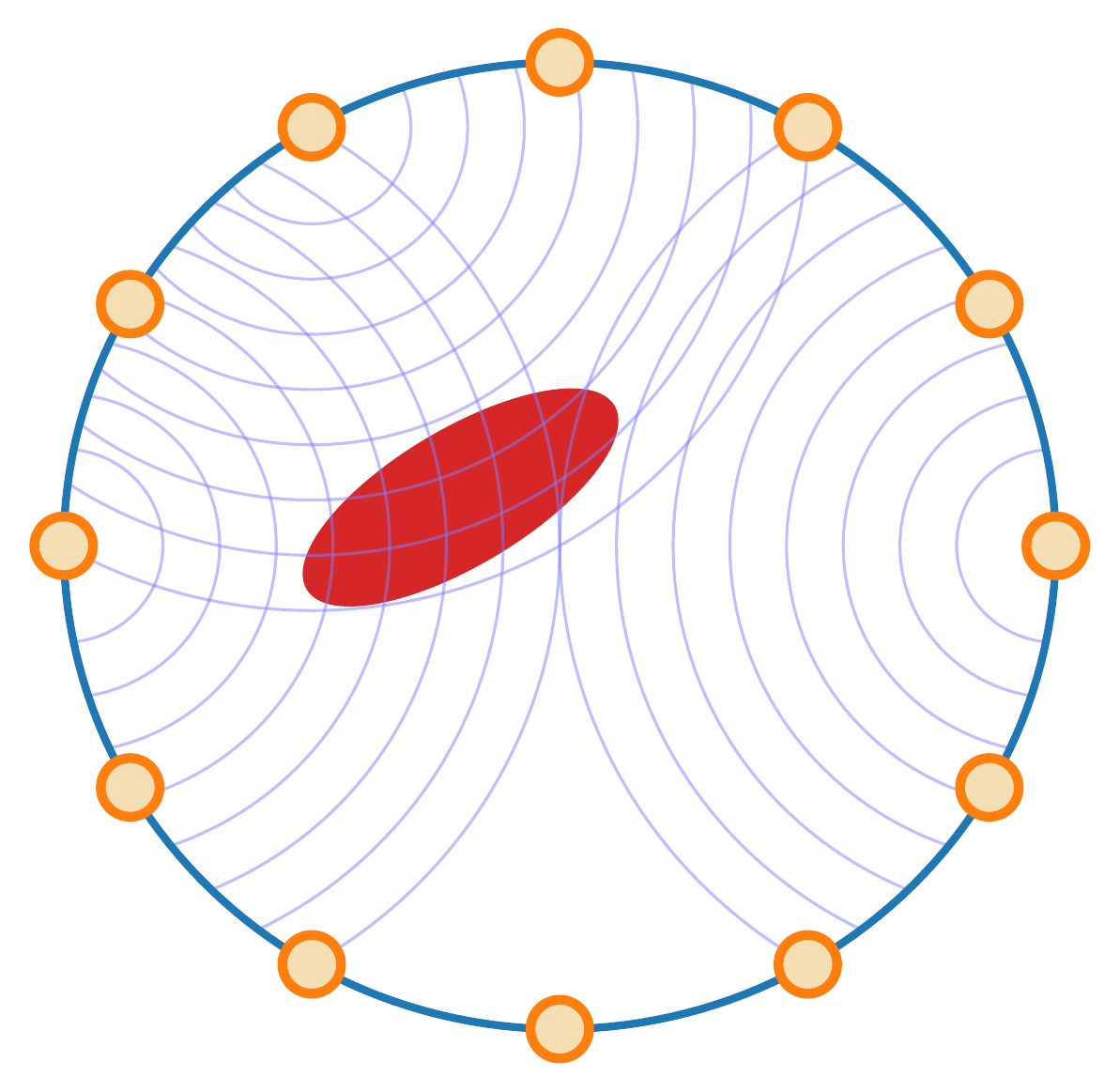}} \hspace{10mm}
\subfloat[Limited view]{\includegraphics[width=0.4\textwidth ]{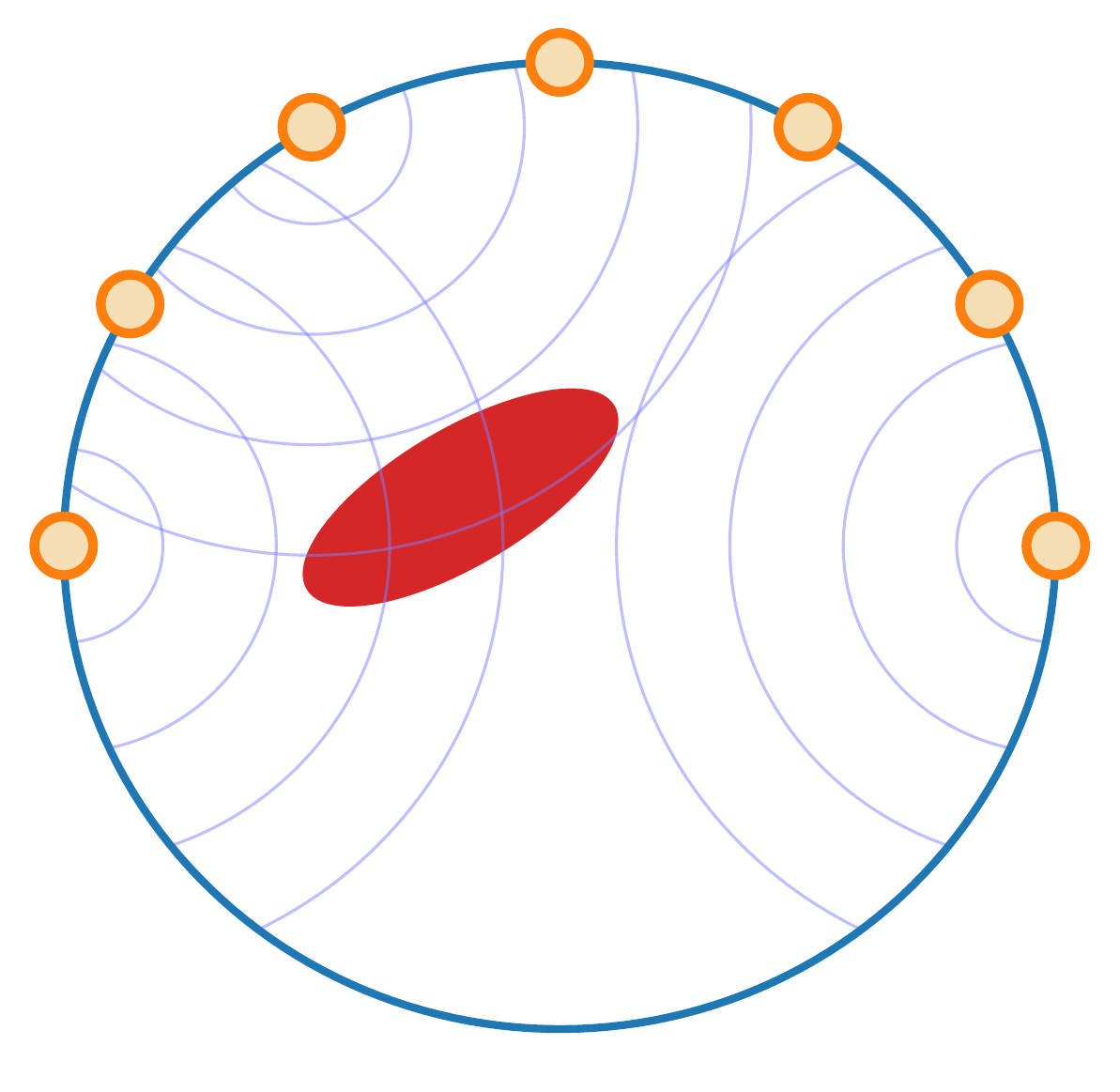}}
\caption{Full and limited view setups}
\label{fig:data_set}
\end{figure}

The PAT inversion problem \eqref{eq:RadonTransform} was first formulated in two-dimensions by \cite{ambar2010}, where the authors derived theoretical conditions on the existence and uniqueness of the solution of \eqref{eq:RadonTransform} when the imaging domain $\Omega$ lies inside, outside, or both inside and outside the acquisition domain $\partial\Omega$. They also derived theoretical inversion formulae for solving \eqref{eq:RadonTransform}. In {\cite{ambar2015}}, the authors considered a setup similar to \eqref{eq:RadonTransform} and derived a formula for the inversion of elliptic Radon transforms in two dimensions, while a formulae for the inversion of spherical Radon transforms in multi-dimensions was derived in {\cite{ambar2018}}. Theoretical inversion formulae for  the circular and spherical Radon transforms with limited angular aperture setups in multi-dimensions was derived in \cite{Syed2016,Gouia2021}. In addition to the theoretical results, numerical algorithms have also been proposed to solve circular and spherical Radon transforms with the partial radii setup. The first numerical algorithm was proposed in \cite{Roy2015} to numerically invert circular and elliptic Radon transforms. Subsequently, numerical algorithms  were proposed for inverting spherical Radon transforms \cite{ambar2018}, and limited angular aperture circular and spherical Radon transforms \cite{Syed2016,Gouia2021}. The aforementioned numerical algorithms were primarily based on a combination of a product trapezoidal rule and truncated singular value decomposition method (TSVD). However, it was observed that the results with these numerical inversion methods were contaminated by several artifacts. Moreover, in a more realistic scenario, $g$ is only known for $\phi \in I \subsetneq [0,2\pi)$ due to the high costs of transducers and possible presence of inaccessible zones for transducer placement, which results in a limited view setup (see Figure \ref{fig:data_set}). In this case, the reconstructions with the TSVD method can suffer from severe artifacts.

Deep learning has proven to be a powerful tool to develop efficient algorithms to solve inverse problems arising in medical imaging (e.g., see \cite{Adler_2017,Kang2018,Adler2018,Vu2020,Schwab2020,Afkham_2021,Celledoni_2021,mukherjee2021endtoend,rudzusika2021invertible}), and overcome computational bottlenecks faced by the more traditional algorithms. In the context of PAT, there are primarily three classes of deep learning-based algorithms to recover the initial pressure distribution. The first is the so called direct approach, where neural networks are trained to directly infer the field from the raw measurement \cite{Waibel2018,Grohl2018}. In \cite{Tong2020}, a two-network reconstruction algorithm was proposed, where the first network transforms the measurement to the image space motivated by the universal back-projection formula (UBP), while the latter network improves the output of the first network. The second, and perhaps the more popular class of approaches comprises training a network to post-process the low-fidelity reconstructions obtained with a cheap (traditional) algorithm, such a filtered back-projection \cite{Antholzer2018,Antholzer2019DeepLF,Davoudi2019,Guan2020FullyDU}. In \cite{Antholzer2018b}, a nullspace projection method was considered to ensure data consistency during the network-based post-processing step. A modified weighted UBP, where the weights are also learnable parameters, was considered in \cite{Schwab2018} to improve the input to the post-processing network. In \cite{Schwab2019}, a TSVD approach was used to obtain the intermediate reconstruction, where the truncation parameter is empirically chosen based on the measurement noise level. Thereafter, a network was trained to recover a stable form of the truncated coefficients. The third class of techniques are the model-based learned iterative reconstructions. In \cite{Hauptmann2018}, a sequence of networks was used to learn the update operators of the proximal gradient descent scheme, with the prior knowledge implicitly learned from the data as opposed to using hand-crafted priors. A partially-learned algorithm was developed in \cite{Boink2020}, where a sequence of networks were trained to approximate the primal and dual update operators, with the algorithm being able to simultaneously reconstruct the image and perform image segmentation. The Network Tikhonov (NETT)-framework was applied to the PAT problem in \cite{Antholzer2021}, where a variational problem was solved with a learned regularizer. It is worth noting that the performance of both post-processing and iterative algorithms depends on the quality of the initial reconstruction, which is typically obtained using a traditional linear reconstruction algorithm. For an in-depth review of existing deep learning algorithms to solve the PAT problem, we refer interested readers to \cite{Hauptmann2020}.

The existing deep learning strategies have been constructed for the full radial setup, while the focus of the present work is on the partial radial problem. Further, to the best of our knowledge, the TSVD method is the only numerical algorithm currently available for this setup, which suffers from the computational challenge of having to pick a problem- and noise-dependent truncation parameter. In the absence of a suitable algorithm to produce a reasonable initial (low resolution) reconstruction for this setup, we propose an end-to-end deep learning algorithm where a U-Net is trained on paired samples to recover the initial pressure field from raw measurements. In other words, our approach falls within the first class of deep learning algorithms discussed above. Note that the input and output data spaces are very different from each other. So the trained U-Net also extracts (from the data) and learns  the underlying (inverse) physical mapping that relates the sinogram to the pressure field reconstruction. As experimental data is presently unavailable for the present setup, we use synthetic phantoms comprising superposition of ellipses to represent the initial field, and generate the corresponding sinogram (measurement) using the forward model. We numerically demonstrate the necessity of training with noisy measurement data to ensure a robust performance of the proposed algorithm. 

The rest of the paper is structured as follows. Section \ref{sec:tsvd} presents a description of the theoretical inversion formula, traditionally used to solve the PAT problems, followed by the TSVD algorithm that serves as the base model for numerical comparison. We describe the proposed deep learning algorithm in Section \ref{sec:dl_method}, with additional details about the network architecture presented in Appendix \ref{app:unet_blocks}. In Section \ref{sec:results}, we present the numerical results of the method when working with both noisy and noise-free measurements, and make concluding remarks in Section \ref{sec:conclusions}. 

\section{Theoretical inversion formula}\label{sec:tsvd}
We expand $f(r,\theta)$ and $g(\rho,\phi)$, appearing in \eqref{eq:RadonTransform}, into a Fourier series:
\[
f(r,\theta)=\NT\NT\NT\sum\limits_{n=-\infty}^{\infty} f_{n}(r)e^{\I n\theta},~ g(\rho,\phi)=\NT\NT\NT\sum\limits_{n=-\infty}^{\infty} g_{n}(\rho)e^{\I n\phi}.
\]
Then, the Fourier coefficients $f_n$ and $g_n$ are related as follows
\Beq \label{eq:Interior-circular-Volterra-integral}
g_{n}(\rho)= \int\limits_{0}^{\rho} \frac{K_{n}(\rho,u)F_{n}(u)}{\sqrt{\rho-u}} \D u,
\Eeq
where
\begin{align}
\notag &F_{n}(u)=f_{n}(R-u), \qquad
T_{n}(x)=\cos(n\arccos(x))\\
\label{eq:Interior-circular-kernel}&K_n(\rho,u) = \frac{4\rho(R-u)T_n\left[\frac{(R-u)^2+R^{2}-\rho^2}{2R(R-u)}\right]}{\sqrt{(u+\rho)(2R+\rho-u)(2R-\rho-u)}}.
\end{align}
Thus, solving for $f(r,\theta)$ is equivalent to solving $f_n$ from $g_n$, using the equation \eqref{eq:Interior-circular-Volterra-integral}, which is a Volterra-type integral equation of the first kind. The following result, given in \cite{ambar2010,ambar2018} (also in \cite{Gouia2021}), guarantees the unique recovery of $f$ from $g$:

\begin{theorem}[Thm. 2.2 \cite{ambar2018}, Thm. 1 \cite{ambar2010}]\label{th:existence}
Let $f(r,\theta)$ be an unknown function, in polar coordinates, supported inside the annulus $A(\delta,R) = \lbrace (r,\theta): r \in (\delta,R),~ \theta \in [0,2\pi]\rbrace$, with $\delta > 0$. If $g(\rho,\phi)$ is known for $\rho \in [0,R-\delta],~ \phi \in [0,2\pi]$, then $f(r,\theta)$ can be uniquely recovered in $A(\delta,R).$ \end{theorem}

To numerically solve for $f$ from \eqref{eq:Interior-circular-Volterra-integral}, a truncated singular value decomposition (TSVD) method, coupled with a product trapezoidal rule, was proposed in \cite{Roy2015}. The method is briefly described as follows: 
\begin{enumerate}
\item We first discretize the domains $[\delta, R-\delta]$ and $[0,2\pi]$, with the corresponding discretized points being $r_i,\rho_i$ and $\theta_j,\phi_j$, respectively. \item For each $r_i,\rho_i$, we consider the discrete Fourier series representations of $f,g$, respectively, using Fast Fourier transforms. 
\item Next, in each subinterval $[\rho_i, \rho_{i+1}]$, we approximate the $F_n(u)K_n(\rho_i,u)$ by a linear function and then integrate with respect to the variable $u$, i.e., we use a product trapezoidal integration step. This removes the integrable singularity $\sqrt{\rho-u}$, in the denominator of \eqref{eq:Interior-circular-Volterra-integral}. 
\item Summing over all subintervals $[0,\rho_i]$, we obtain a matrix equation $ A_n \bf{F}_n = \bf{g}_n$, where $A_n$ is a matrix obtained after the product trapezoidal integration step, $\bf{F}_n$ is a vector of $F_n$, evaluated at each $r_i$, and $\bf{g}_n$ is a vector of $g_n$, evaluated at each $\rho_i$. 
\item Since this matrix turns out to be highly ill-conditioned for several values of $n$, a half-rank TSVD method is employed for obtaining $\bf{F}_n$, from which $f_n$ at each $r_i$ is obtained. 
\item Finally, a discrete inverse Fourier transform is used to recover $f$.
\end{enumerate}

The aforementioned theoretical and numerical inversion formula produces good reconstructions under the assumption that we have a full range of data in the angular variable $\phi$. However, a major drawback of the TSVD method is its inability to provide adequate reconstructions in the case $g$ is not be known for all values of $\phi \in [0,2\pi]$. This is also referred to as a limited view setup. Another drawback of the TSVD method is the presence of large number of artifacts in the case when the number of measurements are highly undersampled in comparison to the function pixel domain. To address these drawbacks, we propose a new machine learning framework for efficient reconstruction of $f$ from its circular Radon transform $g$.

\section{A deep learning-based inversion}\label{sec:dl_method}
\begin{figure}[!htbp]
\centering
\subfloat[Exact phantom]{\includegraphics[width=0.3\textwidth, height=0.3\textwidth]{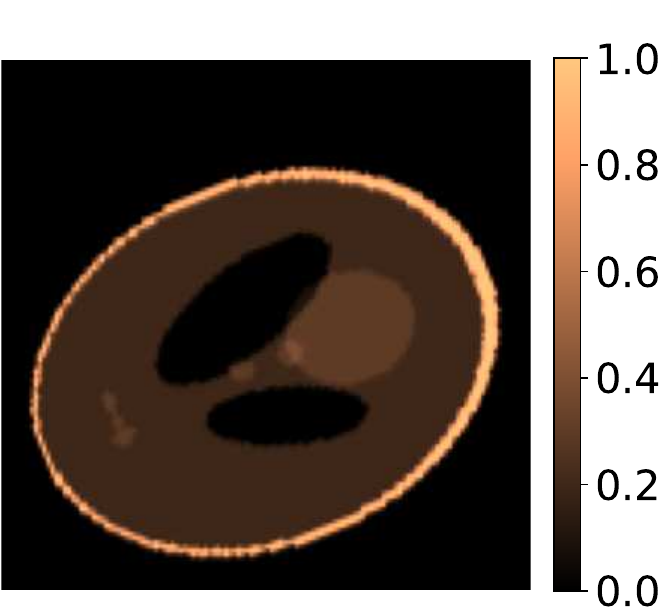}}
\subfloat[$g$]{\includegraphics[width=0.3\textwidth, height=0.3\textwidth]{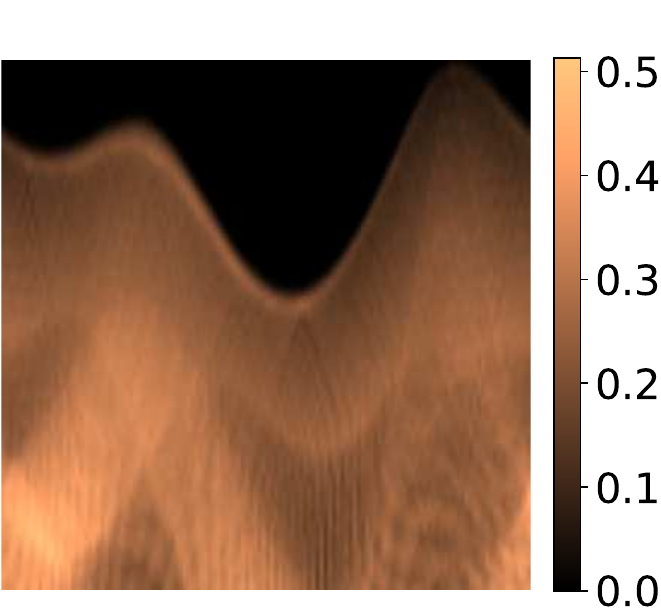}}
\subfloat[Partial view $g$]{\includegraphics[width=0.3\textwidth, height=0.3\textwidth]{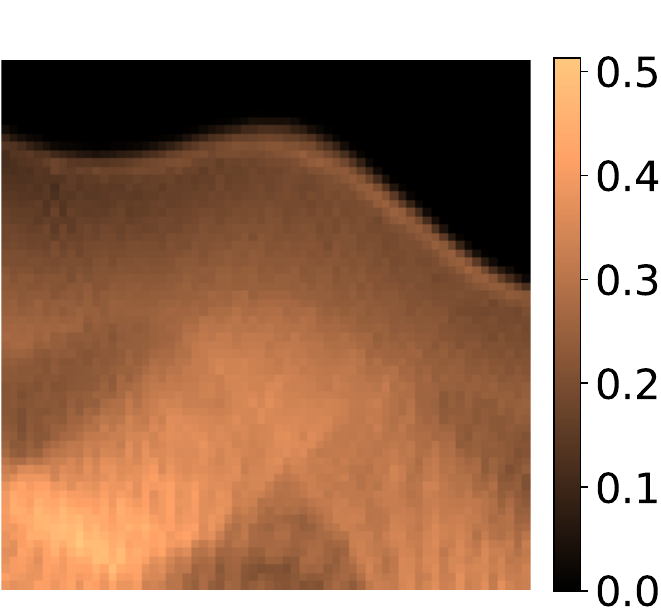}\label{Y1_lim}}\\
\subfloat[15\% Noisy $g$]{\includegraphics[width=0.3\textwidth, height=0.3\textwidth]{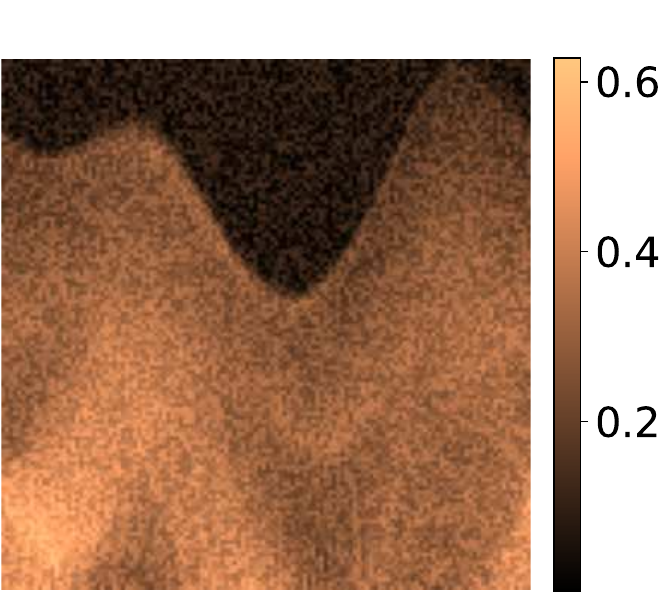}\label{Y1_noisy}}
\hspace{10mm}\subfloat[15\% Noisy and \\ partial view $g$]{\includegraphics[width=0.3\textwidth, height=0.3\textwidth]{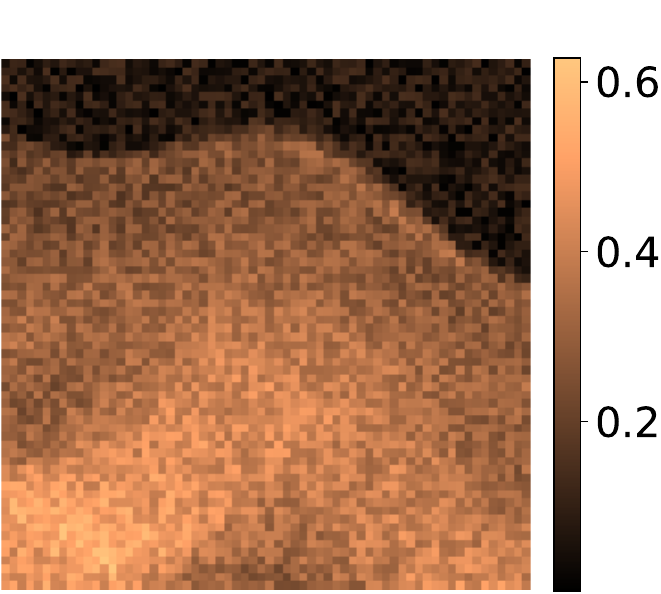}\label{Y1_noisy_lim}}\\
\caption{A test phantom and corresponding measurements. }
\label{fig:phantom}
\end{figure}

We begin by framing the forward problem as
\begin{equation}\label{eqn:forward}
\Y = \F(\X) + \etab
\end{equation}
where $\X$ is the pointwise evaluation of $f$ on a two-dimensional spatial Cartesian grid of size $N_X \times N_X$, $\Y$ is the corresponding circular Radon data on a two-dimensional $\rho-\phi$ Cartesian grid of size $N_Y \times N_Y$, and $\F$ is the forward Radon transform map. $\Y$ might correspond to the measurement corresponding to the full or limited view data (see Figure \ref{fig:data_set}). Further, we assume that the measurement $\Y$ can be corrupted by noise $\etab$ governed by some noise distribution $p_\eta$. Our task is to recover the image $\X$ given an image $\Y$. For the remainder of this paper we set $N_X = 128$, i.e. we fix the resolution of $\X$. When considering a full view measurement, we set $N_Y=128$ with a uniform discretization of $\rho-\phi$ space $[0,1]\times[0,2\pi)$, while for a limited view measurement we set $N_Y=64$ with a uniform discretization of space $[0,1]\times[0,\pi)$. Note that compared to the full view measurement, the limited view measurement spans only half $\phi$ space and performs a coarser sampling (by a factor of 2) in the $\rho$ space. These are depicted in Figure \ref{fig:phantom} for a particular $f$ corresponding to a perturbed Shepp-Logan phantom. 

\subsection{U-Net architecture}
\begin{figure}[!htbp]
\centering
\subfloat[U-Net for full view measurement]{\includegraphics[height=0.4\textwidth]{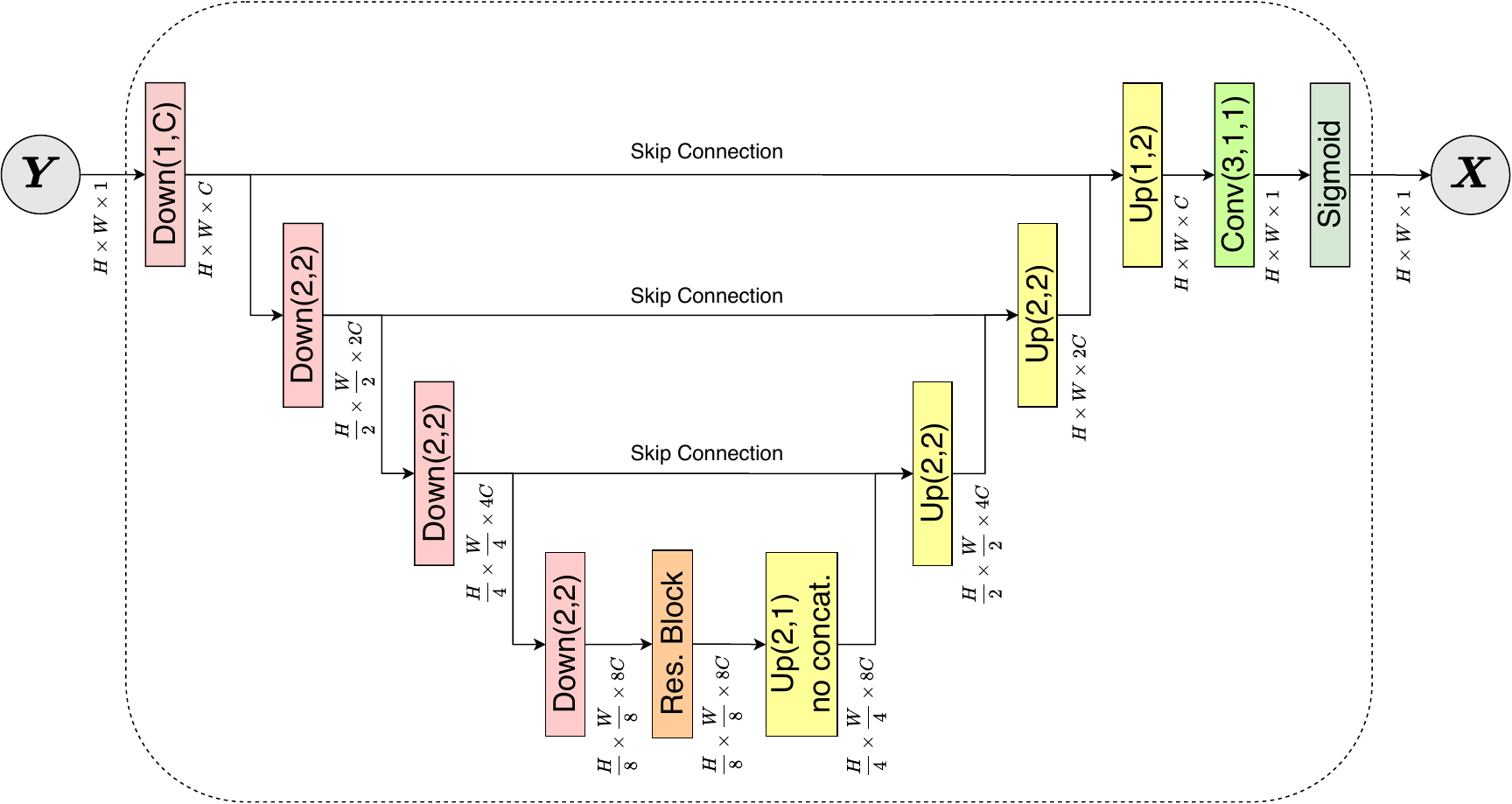}}\\
\subfloat[U-Net for limited view measurement]{\includegraphics[height=0.4\textwidth]{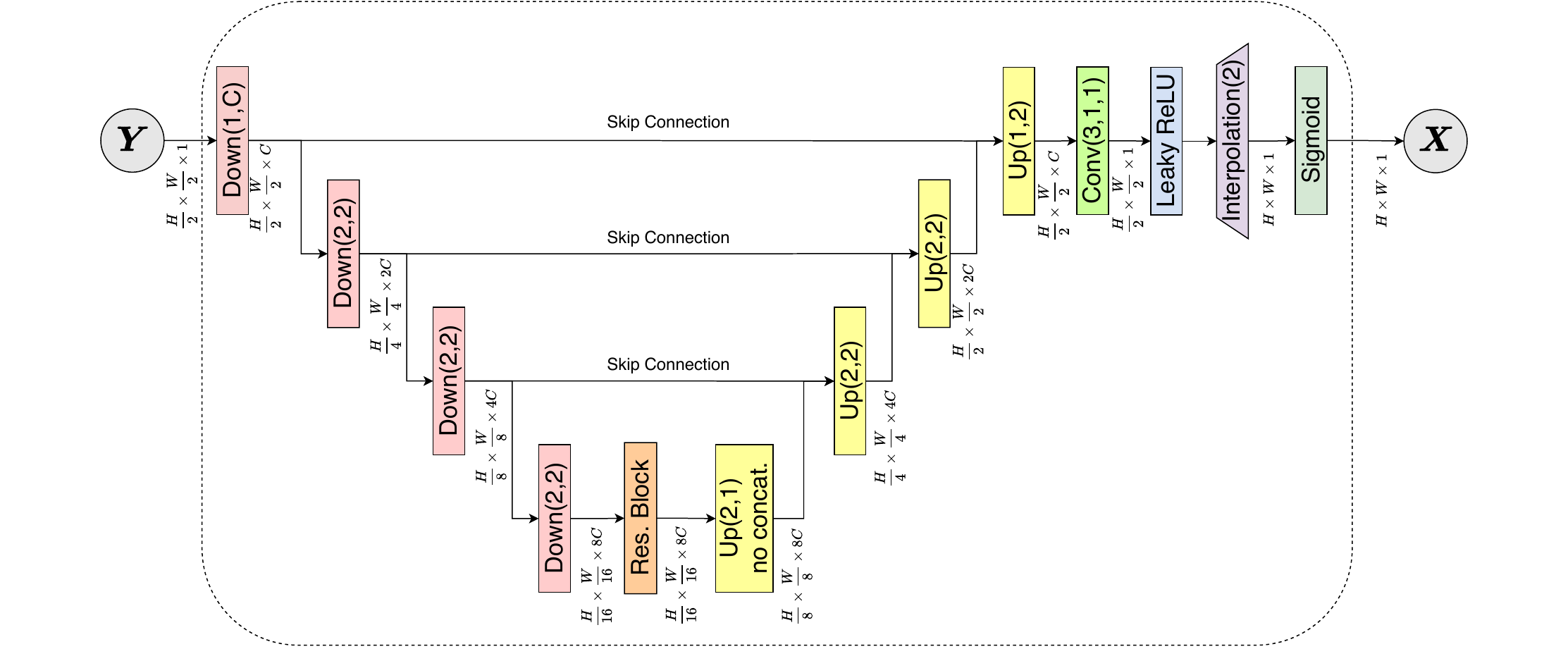}}
\caption{U-Net architectures considered for the inverse problem. We set $H=W=N_X$} and $C=32$.
\label{fig:unet}
\end{figure}

We denote the network by the tensor-valued function
\begin{equation}\label{eqn:unet}
\net(.;\bpsi):\Rb^{N_Y \times N_Y} \rightarrow \Rb^{N_X \times N_X}, \quad \Xt = \net(\Y;\bpsi)
\end{equation}
that is parametrized by the set of trainable weight and biases $\bpsi$. We consider a U-Net architecture, which is commonly used for image-to-image deep learning-based transformations \cite{UNet}. As shown in Figure \ref{fig:unet}, the U-Net takes the input measurement $\Y$ and pushes it down a contraction branch. As it moves through the various levels of this branch, the spatial resolution of the input field decreases while its multi-scale features are extracted and augmented as additional channels (the third tensor dimension shown in Figure \ref{fig:unet}). This is followed by an expanding branch which gradually increases the spatial resolution while reducing the number of extracted features (channels), till the U-Net prediction $\Xt$ of the desired shape is recovered. Further, through skip connections, the U-Net combines the multi-scale features learned in contracting branch with the features learned (at the same spatial resolution) in the expanding branch. We set $H=W=N_X$ and $C=32$ (see Figure \ref{fig:unet}) for all networks trained in the present work. Additional details about the various constitutive blocks of our U-Nets are given in Appendix \ref{app:unet_blocks}.

Note that the U-Net architecture for the full view measurement (Figure \ref{fig:unet}(a)) and limited view measurement (Figure \ref{fig:unet}(b)) are essentially the same, with the only distinction being that the latter takes as input a measurement with half the resolution and uses an additional activation and {\tt Interpolation(2)} layer towards the end to ensure the correct (predefined) output resolution. 

\subsection{Dataset and training}
We train the U-Nets using a supervised learning strategy. The dataset of pairwise samples used to train and test the networks is generated by considering realizations of the perturbed Shepp-Logan (discrete) phantom $\X$ and then evaluating the (noisy) measurement $\Y$ using the forward map \eqref{eqn:forward}. The Shepp-Logan phantom consists of ten ellipses with centres at $(c_k,d_k)$, semi-axis lengths $(a_k,b_k)$, angle of inclination $\psi_k$, and intensity $I_k$. The values of these coefficients can be obtained from \cite{toft_1996}. To perturb the Shepp-Logan phantom, the aforementioned coefficients are modified as follows:
\begin{equation}\label{eq:coef}
\begin{aligned}
&\bar{c}_k = c_k + 0.01\sigma_k^1,~ \bar{d}_k = d_k + 0.01\sigma_k^2,\\
&\bar{a}_k = a_k + 0.01\sigma_k^3,~ \bar{b}_k = b_k + 0.01\sigma_k^4,\\
&\bar{\psi}_k = \psi_k + 0.08\sigma_k^5,~ \bar{I}_k = I_k + 0.001\sigma_k^6,\\
\end{aligned}
\end{equation}
where $\sigma_k^i\sim \mathcal{U}[-0.5,0.5]$. Following \cite{Patel2021}, the intensity of the perturbed Shepp-Logan is scaled to preserve the values between 0 and 1. We then obtain a discrete phantom $\X$ for a single training, test, or validation sample, using the coefficients of the perturbed Shepp-Logan in \eqref{eq:coef} on a grid of size $N_X \times N_X$, and further transforming the discrete phantom. The transformation of the discrete phantom is combination of rotation by an angle $\alpha \sim \mathcal{U}[-0.5,0.5]$ and translation by $n,m$ pixels in the horizontal and vertical direction, respectively, with $n,m\sim \mathcal{U}[-15, -14, \cdots, 14, 15]$.

We generate several types of dataset depending on $N_Y$ and the amount of additive Gaussian noise added to the measurement. The Gaussian noise is given by a normal distribution $N(0,\sigma)$, where $100\sigma^2$ is the noise level percentage.  These are listed in Table \ref{tab:datasets}. The datasets are named using the nomenclature ``\{PHASE\}\{$N_Y$\}\{NOISY-TYPE\}'', where PHASE represents when the dataset is used, $N_Y$ denotes the measurement resolution, while NOISE-TYPE shows the type of noise added. For instance, Train128cn15 corresponds to the dataset used in the training phase with measurement resolution of $128 \times 128$, comprising both clean measurements (c) and noisy measurements (n) with 15\% noise. On the other hand, Test64n5 corresponds to the dataset used in the testing phase with measurement resolution of $64 \times 64$, comprising only noisy measurements with 5\% noise. Note that the datasets used in the training phase are split into training and validation samples. Further, for datasets containing both clean and noisy samples, care has been taken to ensure that an equal number of samples of each type is included. 

\begin{table}[!htbp]
\renewcommand{\arraystretch}{1.5}
\centering
\caption{Datasets used for training, valiation and testing}
\begin{tabular}{l c c l}
\toprule
Dataset & $N_Y$ & \% noise & \begin{tabular}[c]{@{}c@{}}\# samples \end{tabular}\\
\midrule
Train128c      & 128   & 0        & Training: 2000, Validation: 500  \\
Train128cn15   & 128   & 0 \& 15  & Training: 2000, Validation: 500  \\
Test128c      & 128   & 0        & Testing: 500  \\
Test128n5     & 128   & 5        & Testing: 500  \\
Test128n15    & 128   & 15        & Testing: 500  \\
Train64cn15    & 64    & 0 \& 15  & Training: 2000, Validation: 500  \\
Test64n5      & 64   & 5        & Testing: 500  \\
Test64n15     & 64   & 15        & Testing: 500  \\
\bottomrule
\end{tabular}\label{tab:datasets}
\end{table}

Training the network involves prescribing $\bpsi$ to ensure that the predictions $\Xt$ are close to the ground truth $\X$ for all $(\X,\Y)$ pairs in the training set. The objective function is taken to be the mean-squared error loss function
\begin{equation}\label{eqn:loss}
    \mathcal{L}(\bpsi) = \frac{1}{N_b N_X^2} \sum_{i=1}^{N_b} \|\X^{(i)} - \net(\Y^{(i)};\bpsi)\|^2,
\end{equation}
where $\|.\|$ is the Euclidean norm and $N_b$ is size of the mini-batch randomly sampled from the training set. We then solve the minimization problem
\begin{equation}\label{eqn:minimize}
\bpsi^* = \argmin{\bpsi} \ \mathcal{L}(\bpsi),
\end{equation}
using a stochastic gradient descent type algorithm, with the final trained U-Net given by $\net(.;\bpsi^*)$. The U-Nets in the present work are trained using the Adam optimizer \cite{kingma2017adam} with a learning rate of $1.0e-3$, $\beta_1=0.5$ and $\beta_2 = 0.9$. We use a batch size $N_b = 100$, set the Leaky ReLU activation parameter as $\alpha = 0.1$. Additionally, we augment the loss function with an $L^2$ regularization of the network weights, with a regularization parameter $1e-5$. The networks are trained for 300 epochs, where one epoch corresponds to a full sweep over the entire training set. Our computations were performed using Tensorflow2 on an iMac Pro (2017) with a 3 GHz 10-Core Intel Xeon W processor and 64 GB 2666 MHz DDR4 RAM. The time taken for the entire training and validation process was approximately 46 hours.

\section{Numerical results}\label{sec:results}
We present numerical results to demonstrate the performance of the proposed deep learning approach for the PAT inversion problem. We first train an appropriate U-Net that leads to accurate reconstructions for the partial radial but full view setup, i.e., $N_Y = 128$. This network needs to be robust to measurement noise. Based on the findings of the full view setup, we then proceed to train a U-Net for limited view measurements. We also compare the U-Net reconstructions with those obtained with the traditional TSVD inversion approach outlined in Section \ref{sec:tsvd}

\begin{figure}[!htbp]
\centering
\subfloat[UNet128c]{\includegraphics[width=0.3\textwidth, ]{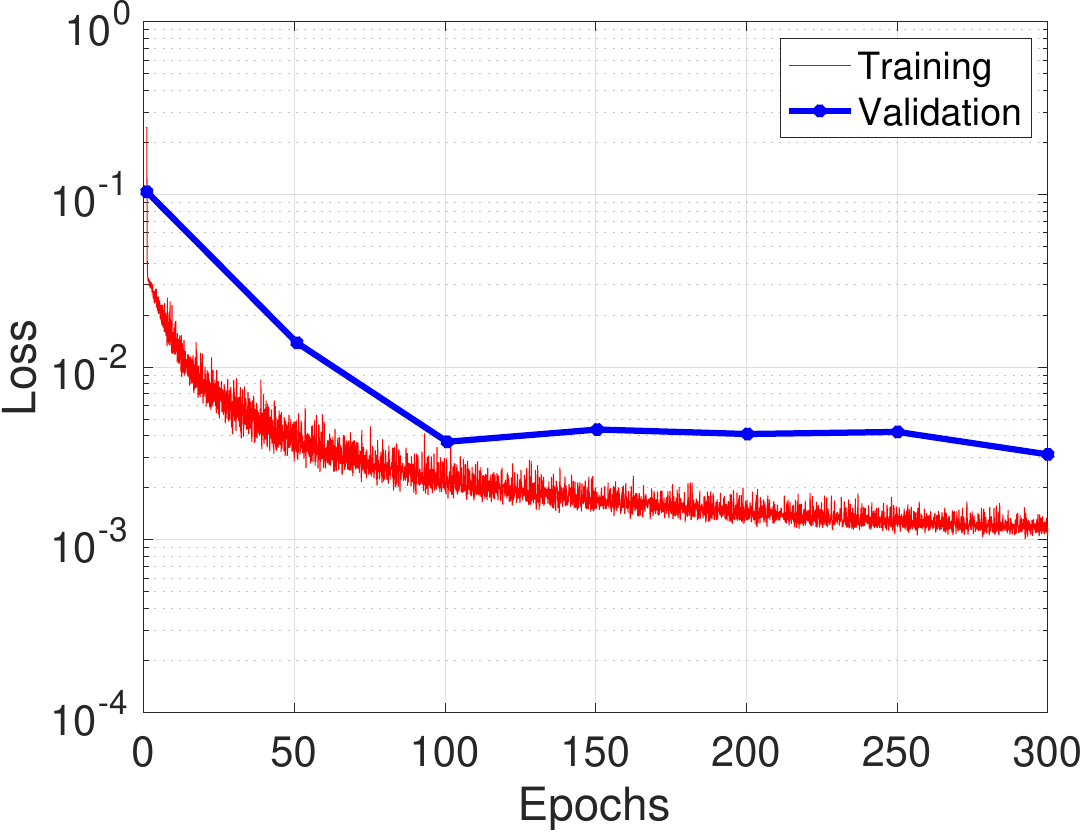}\label{trainNetA}} 
\subfloat[UNet128cn15]{\includegraphics[width=0.3\textwidth, ]{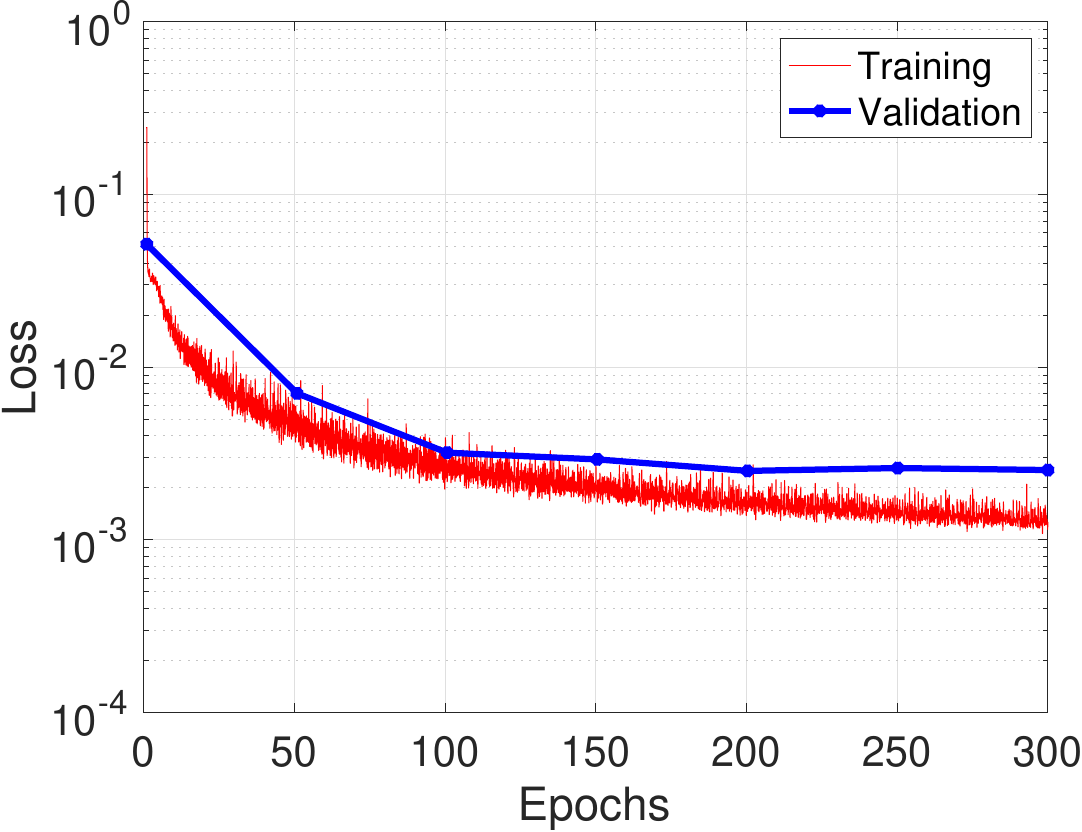}\label{trainNetB}}
\subfloat[UNet64cn15]{\includegraphics[width=0.3\textwidth, ]{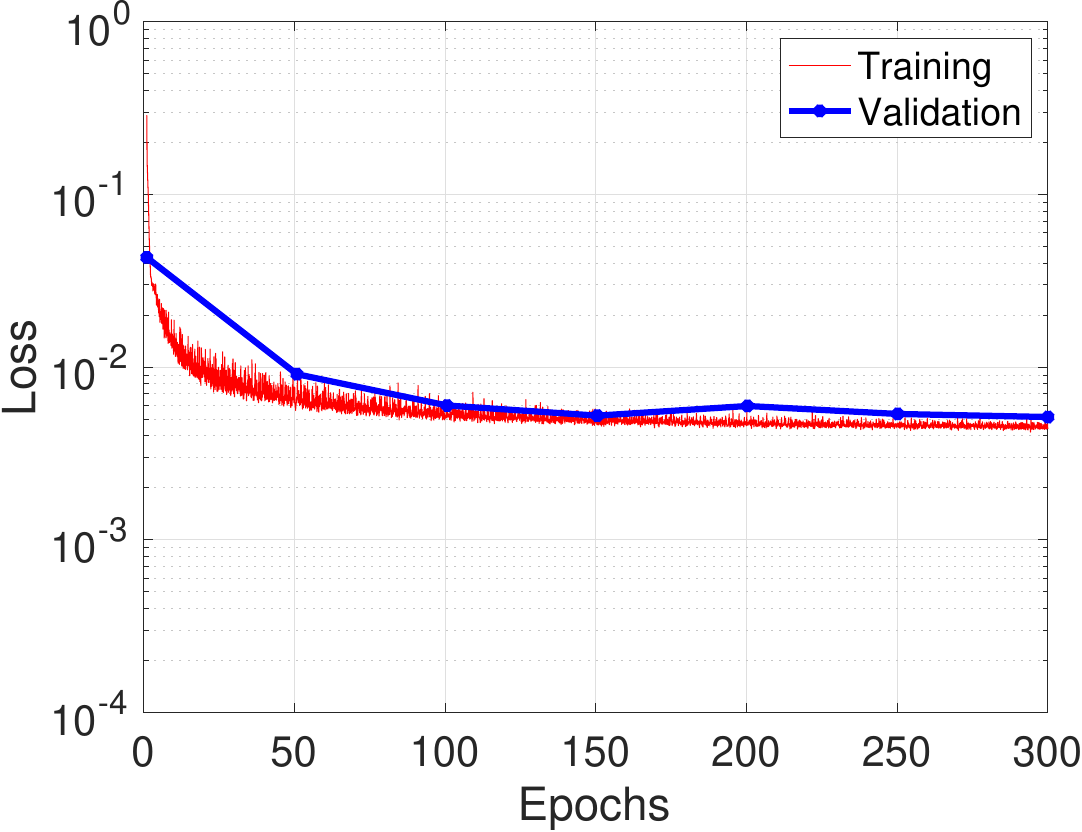}\label{trainNetC}}
\caption{Training and validation losses.}
\label{fig:losses}
  \end{figure}

\subsection{Results for full view setup}
\begin{figure}[!htbp]
\centering
\begin{tabular}{ m{0.2cm} >{\centering}m{3.2cm} >{\centering}m{3.2cm} >{\centering}m{3.2cm} }
& \textbf{Test Sample 1} & \textbf{Test Sample 2} & \textbf{Test Sample 3} \tabularnewline
\rotatebox{90}{\textbf{True $\X$}} & \includegraphics[width=0.29\textwidth]{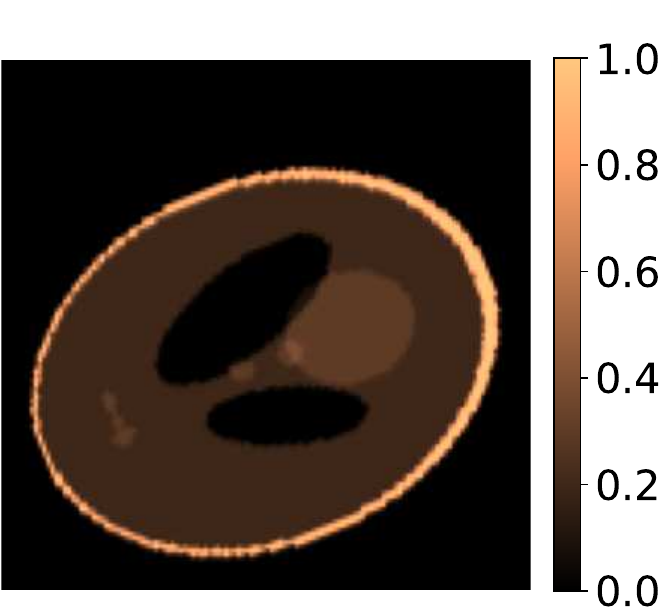} & \includegraphics[width=0.29\textwidth]{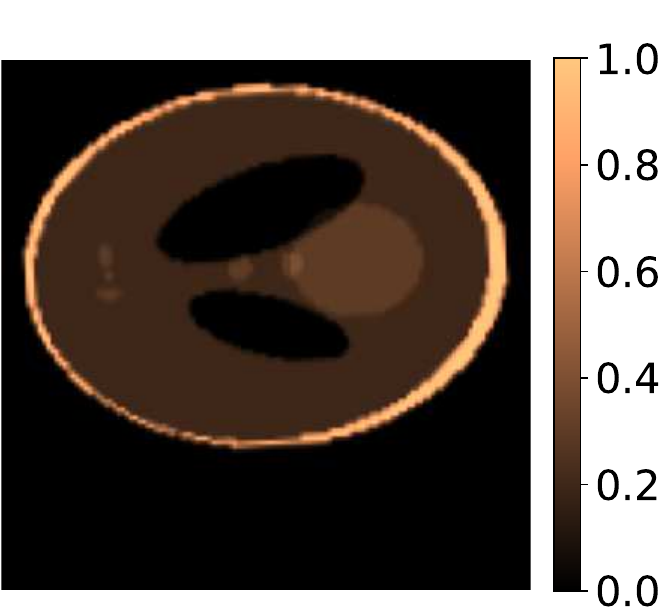}& \includegraphics[width=0.29\textwidth]{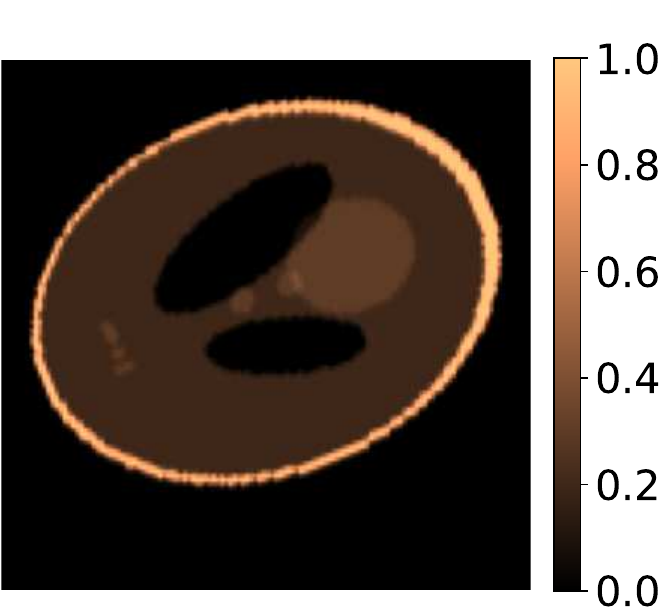} \tabularnewline
\rotatebox{90}{\textbf{Measurement}} & \includegraphics[width=0.29\textwidth]{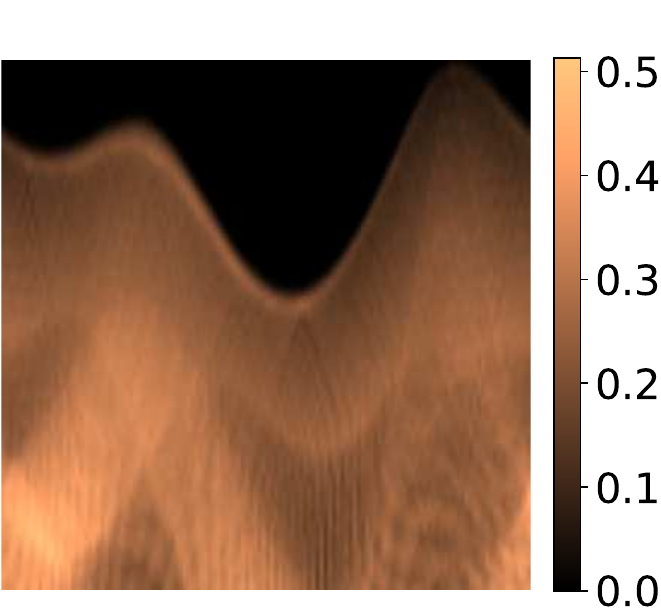} & \includegraphics[width=0.29\textwidth]{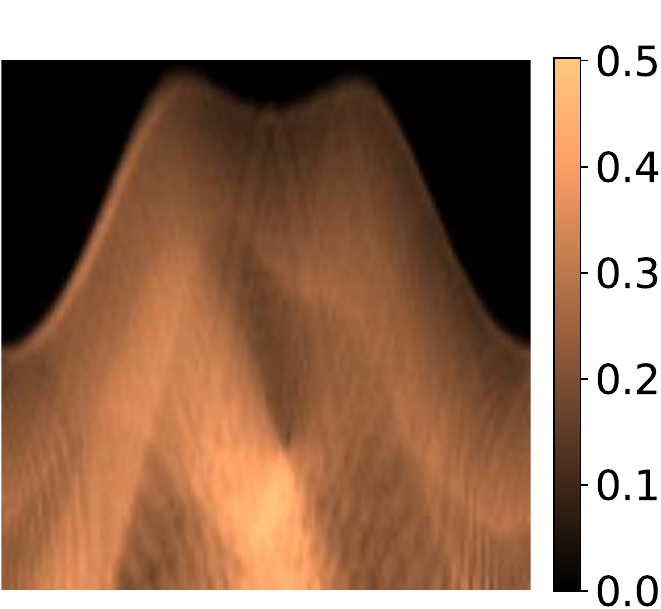}& \includegraphics[width=0.29\textwidth]{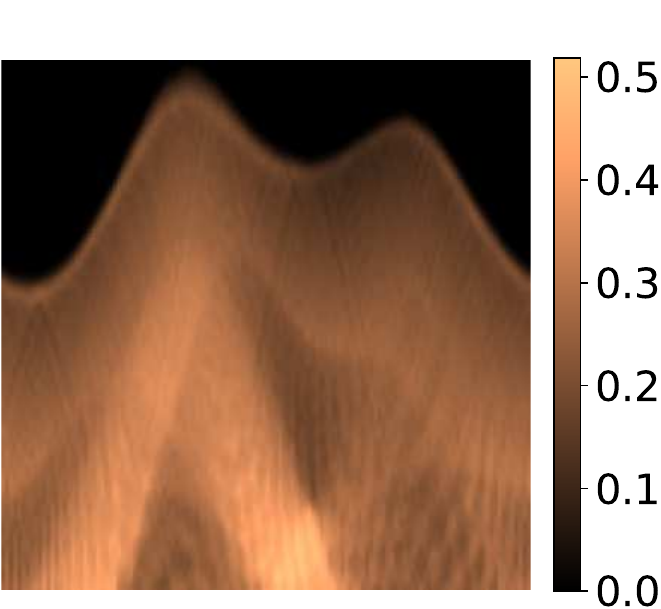} \tabularnewline
\rotatebox{90}{\textbf{UNet128c}} & \includegraphics[width=0.29\textwidth]{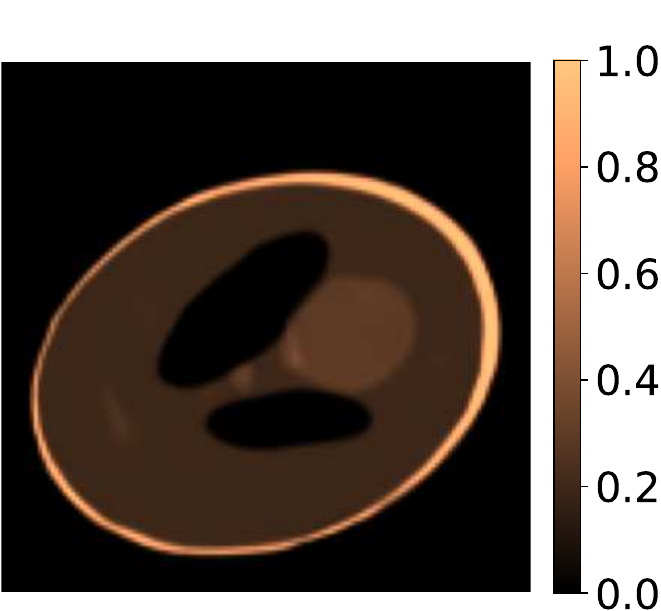} & \includegraphics[width=0.29\textwidth]{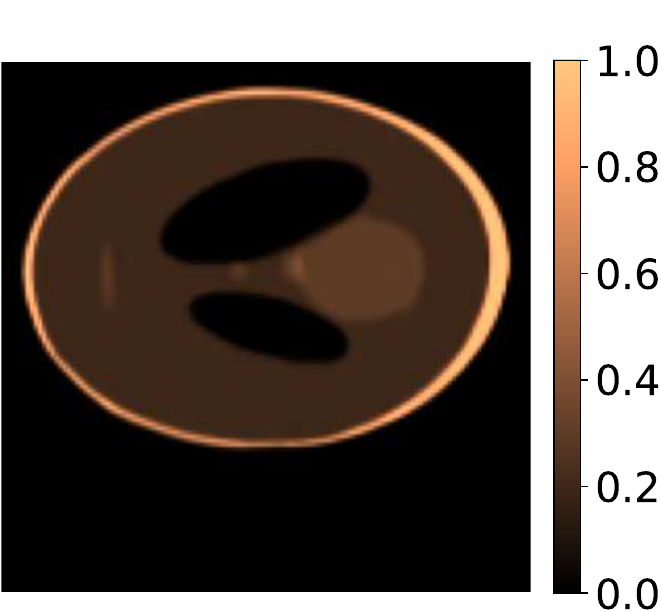}& \includegraphics[width=0.29\textwidth]{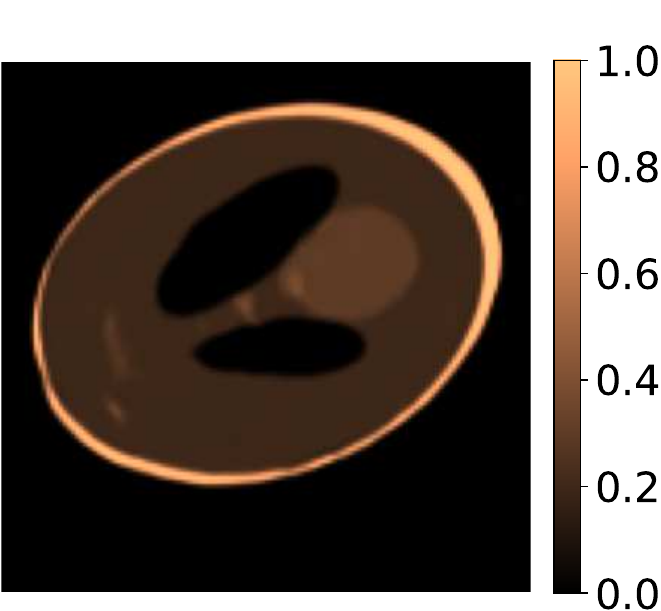} \tabularnewline
\rotatebox{90}{\textbf{TSVD}} & \includegraphics[width=0.28\textwidth]{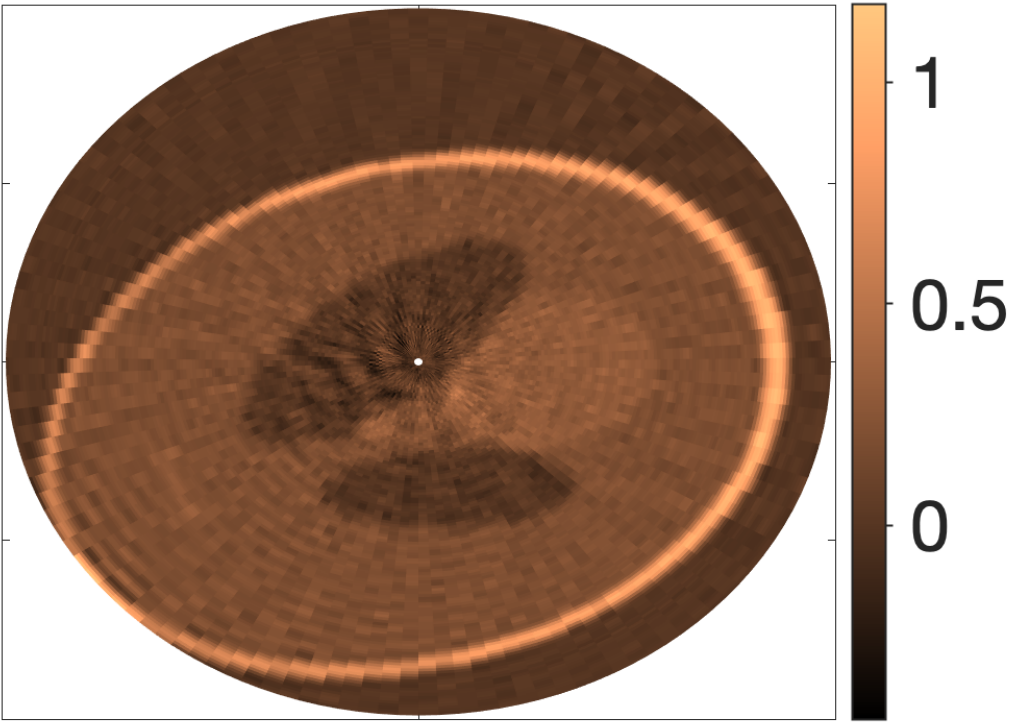} & \includegraphics[width=0.28\textwidth]{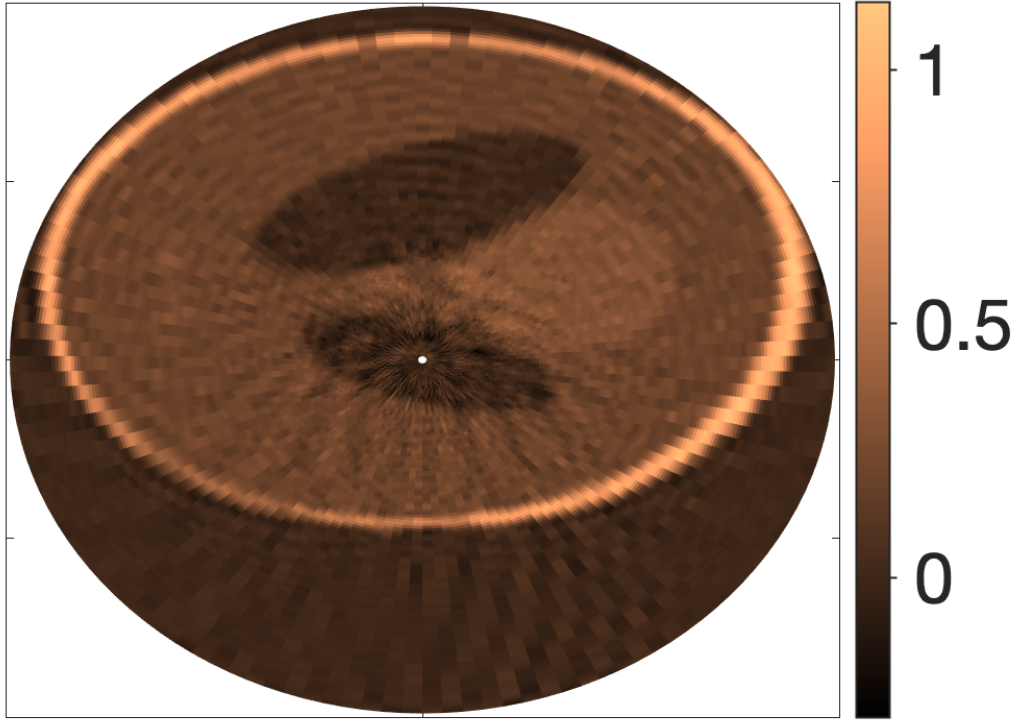}& \includegraphics[width=0.28\textwidth]{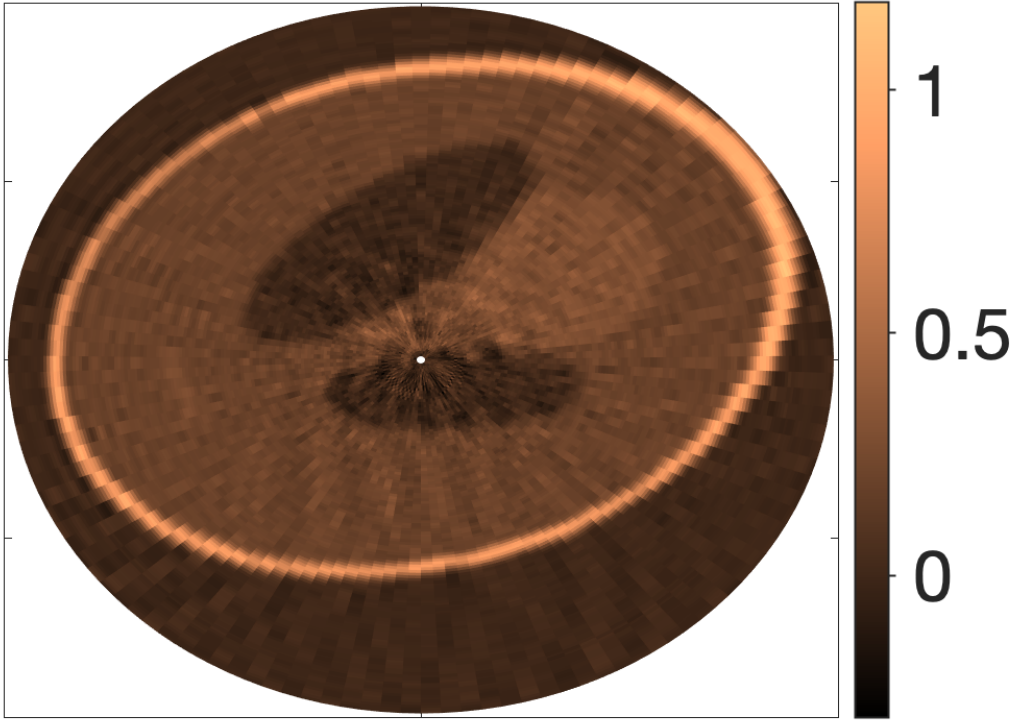}
\end{tabular}
\caption{Results for samples from test set Test128c with UNet128c and TSVD. }
\label{fig:NetA_QI}
\end{figure}

We consider a U-Net trained on the dataset Train128c, which comprises noise free full view measurements. We term the network as UNet128c, with its training and validation losses (as a function of the number of epochs) shown in Figure \ref{fig:losses}(a). Note that  the validation loss is evaluated after intervals of 50 epochs. The trained network is tested on Test128c comprising the same type of samples used to train the network. As shown in Figure \ref{fig:NetA_QI}, the network is able to reconstruct the phantom from the measurement very accurately. Furthermore, unlike the TSVD algorithm, the reconstructions with the network are not contaminated by streak artefacts. 

\begin{figure}[!htbp]
\centering
\begin{tabular}{ m{0.2cm} >{\centering}m{3.2cm} >{\centering}m{3.2cm} >{\centering}m{3.2cm} }
& \textbf{Test Sample 1} & \textbf{Test Sample 2} & \textbf{Test Sample 3} \tabularnewline
\rotatebox{90}{\textbf{True $\X$}} & \includegraphics[width=0.29\textwidth]{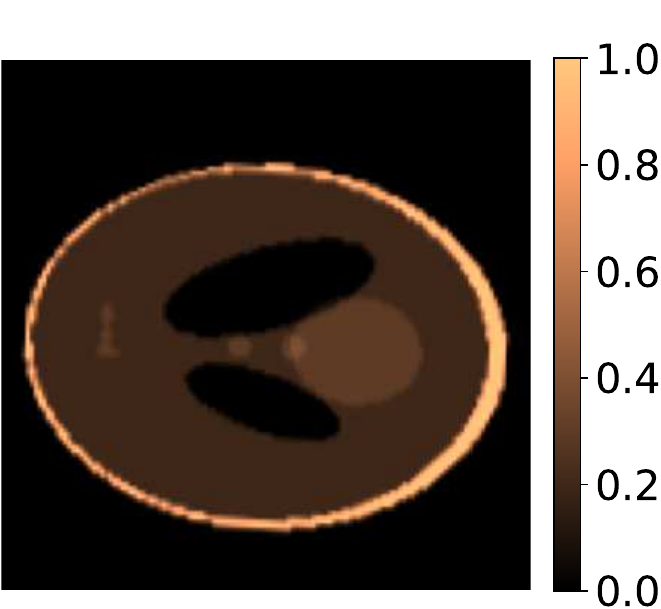} & \includegraphics[width=0.29\textwidth]{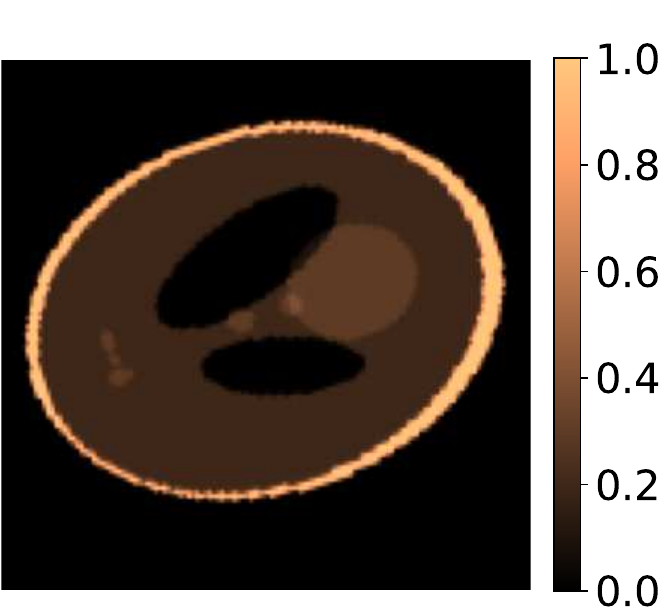}& \includegraphics[width=0.29\textwidth]{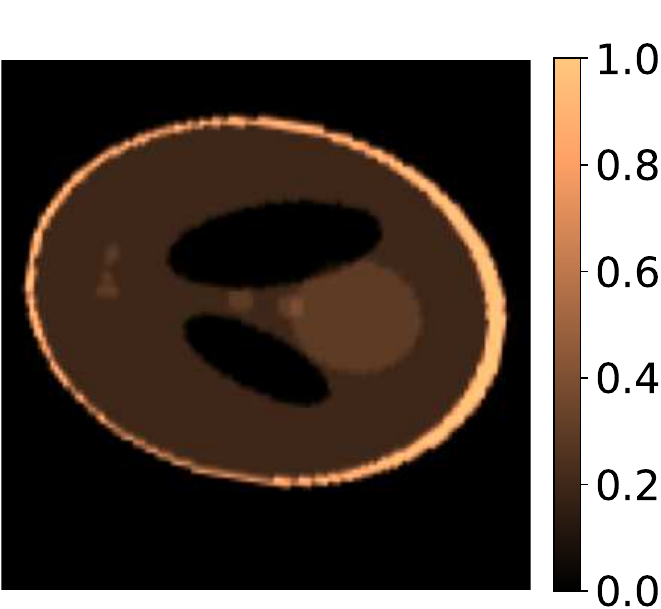} \tabularnewline
\rotatebox{90}{\textbf{Measurement}} & \includegraphics[width=0.29\textwidth]{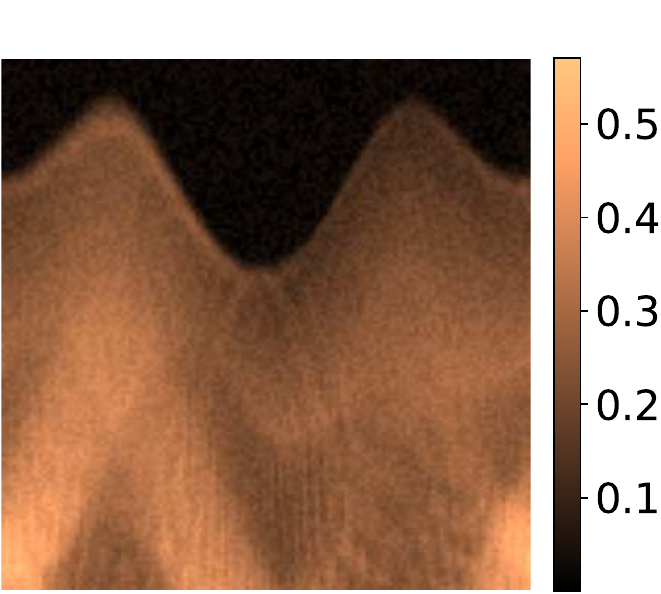} & \includegraphics[width=0.29\textwidth]{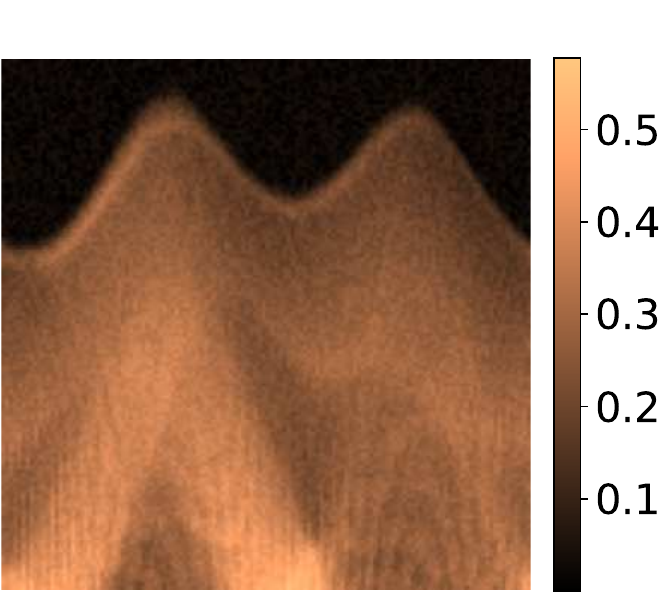}& \includegraphics[width=0.29\textwidth]{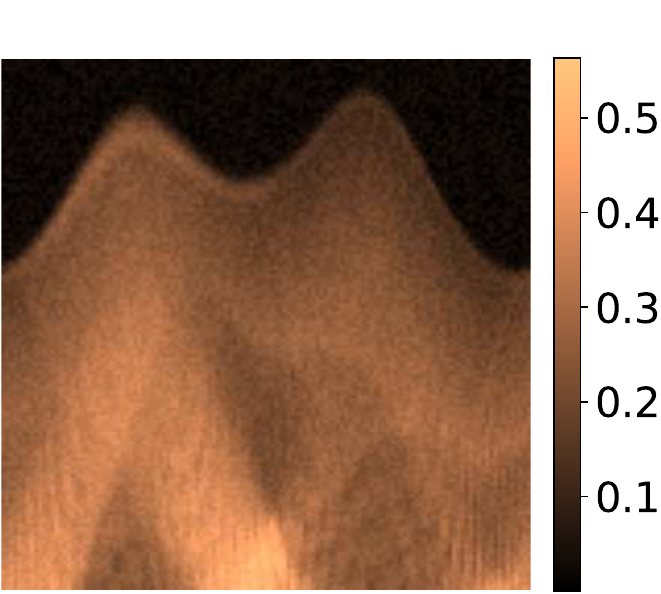} \tabularnewline
\rotatebox{90}{\textbf{UNet128c}} & \includegraphics[width=0.29\textwidth]{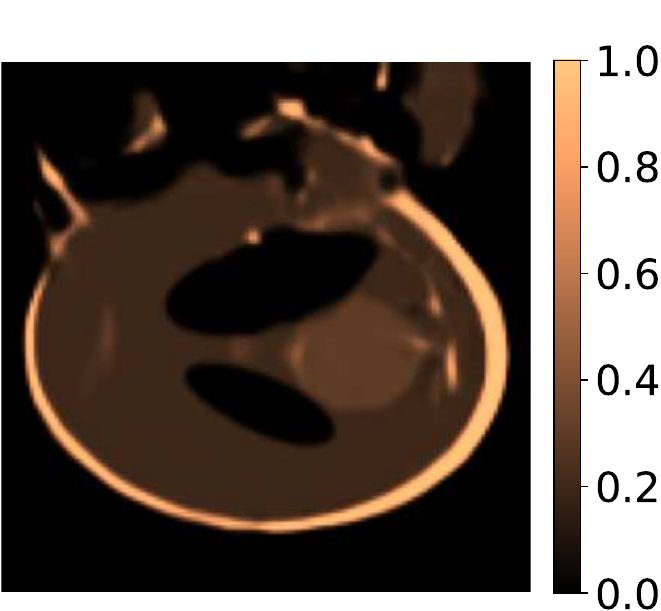} & \includegraphics[width=0.29\textwidth]{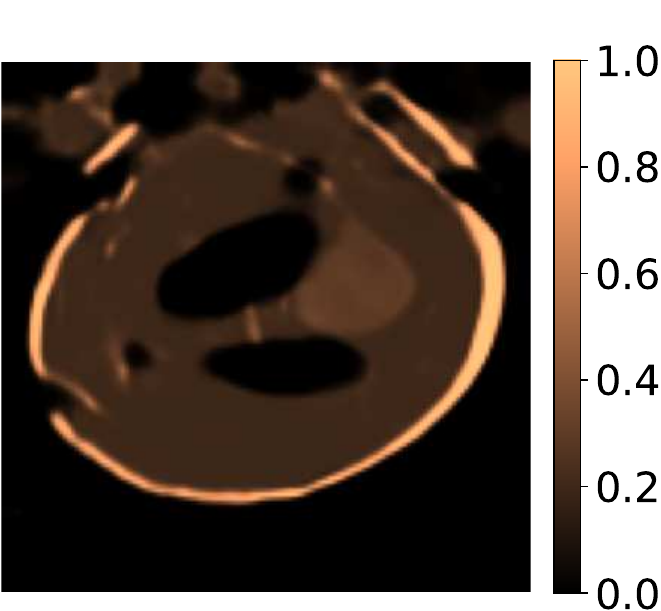}& \includegraphics[width=0.29\textwidth]{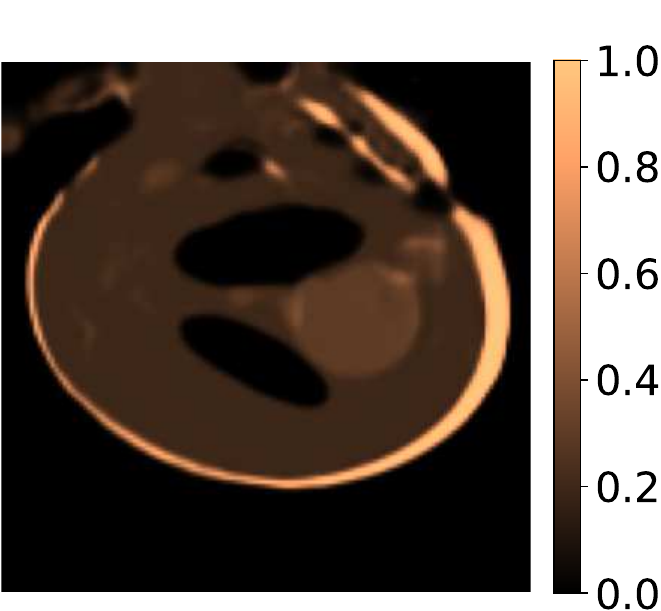} \tabularnewline
\rotatebox{90}{\textbf{TSVD}} & \includegraphics[width=0.295\textwidth]{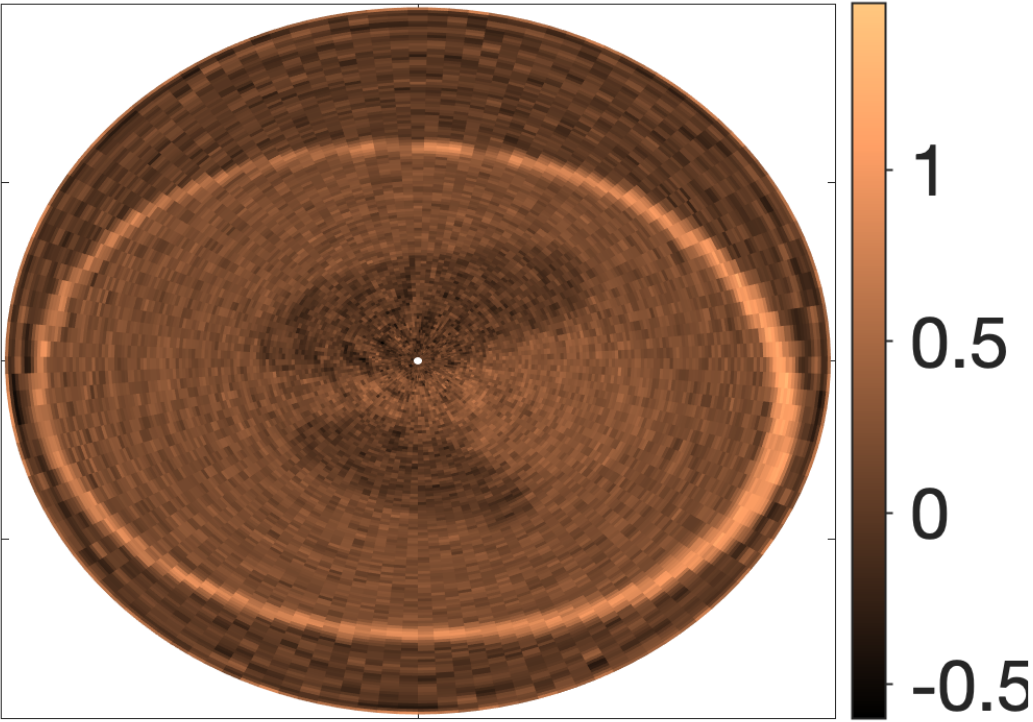} & \includegraphics[width=0.295\textwidth]{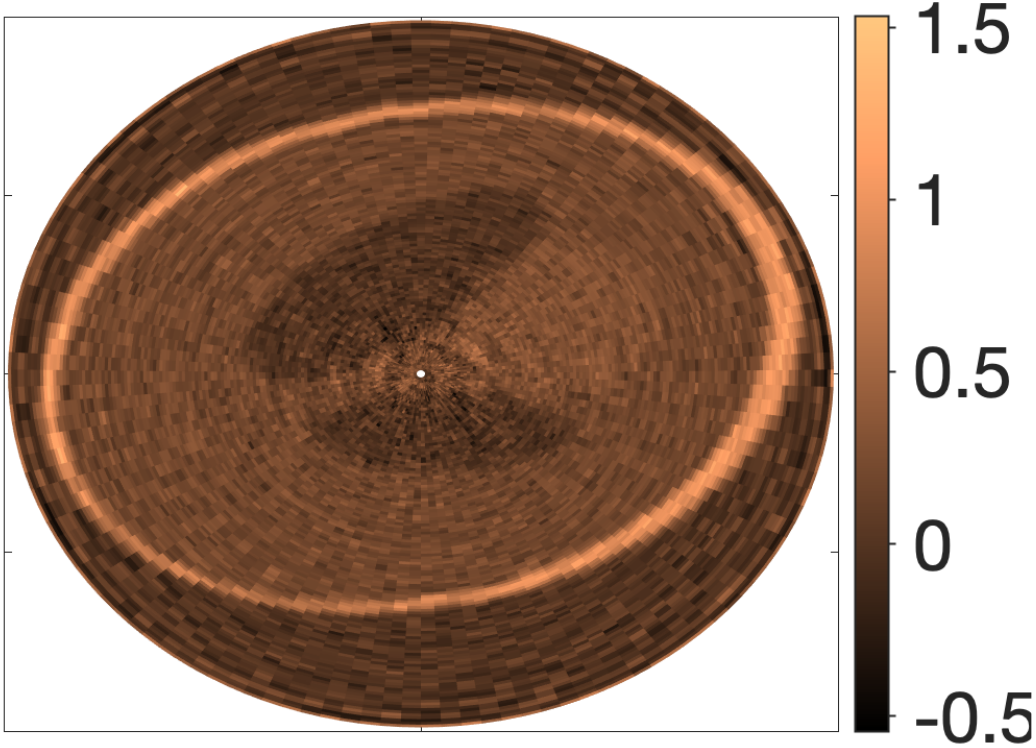}& \includegraphics[width=0.295\textwidth]{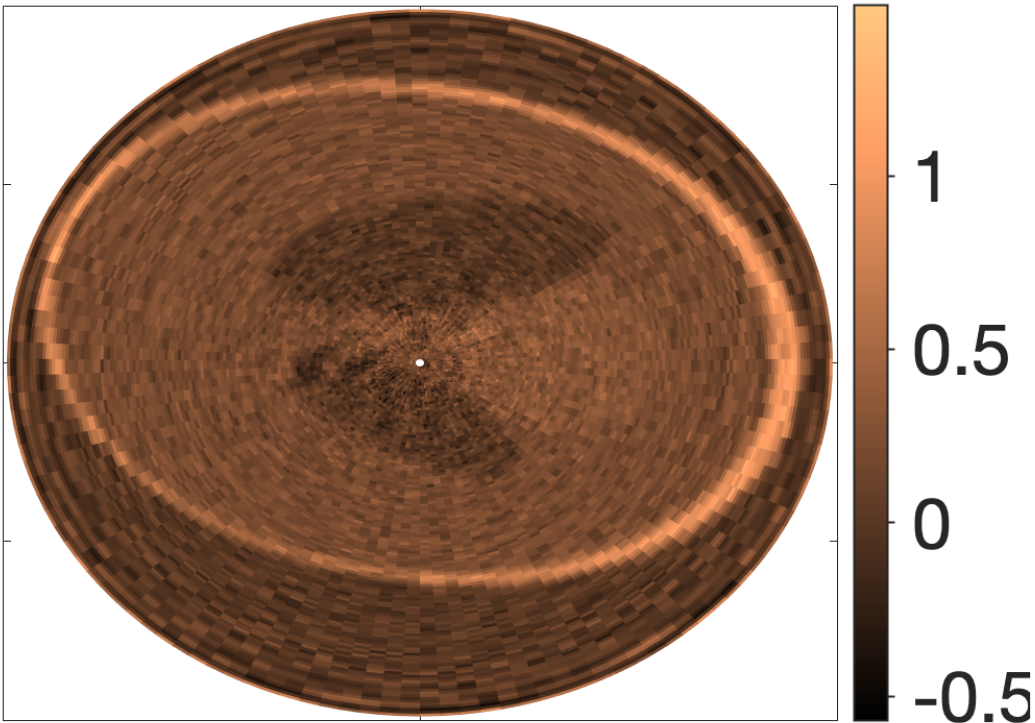}
\end{tabular}
\caption{Results for samples from test set Test128n5 with UNet128c and TSVD. }
\label{fig:NetA_QII}
\end{figure}

\begin{figure}[!htbp]
\centering
\begin{tabular}{ m{0.2cm} >{\centering}m{3.2cm} >{\centering}m{3.2cm} >{\centering}m{3.2cm} }
& \textbf{Test Sample 1} & \textbf{Test Sample 2} & \textbf{Test Sample 3} \tabularnewline
\rotatebox{90}{\textbf{True $\X$}} & \includegraphics[width=0.29\textwidth]{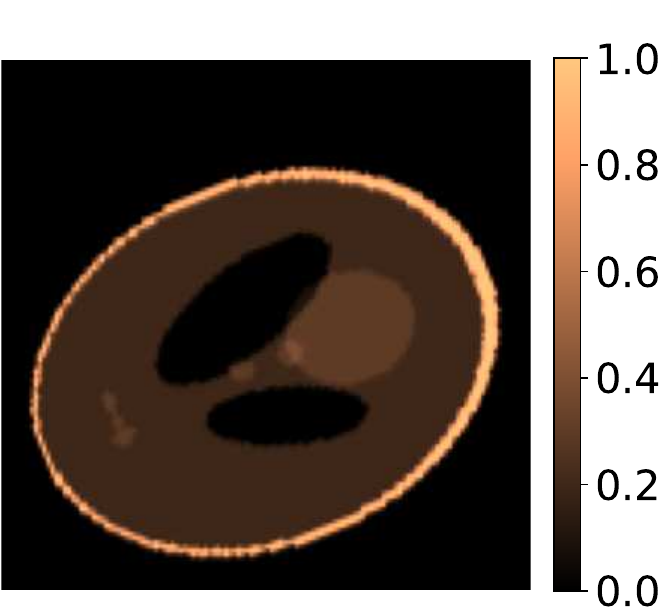} & \includegraphics[width=0.29\textwidth]{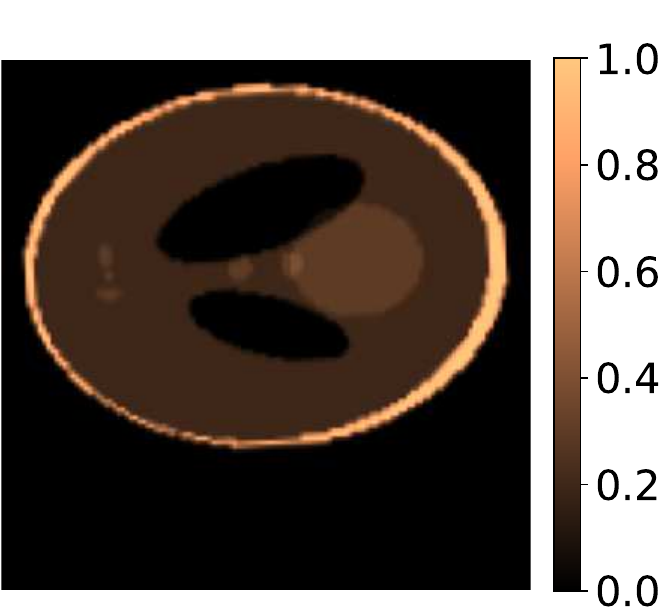}& \includegraphics[width=0.29\textwidth]{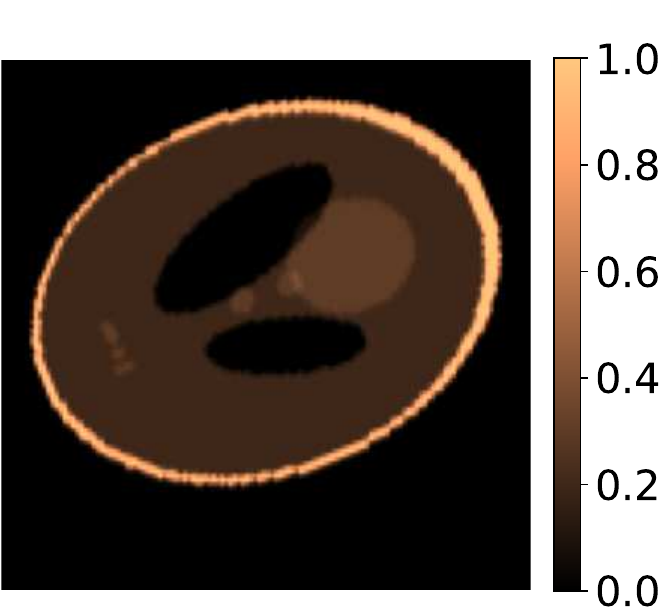} \tabularnewline
\rotatebox{90}{\textbf{Measurement}} & \includegraphics[width=0.29\textwidth]{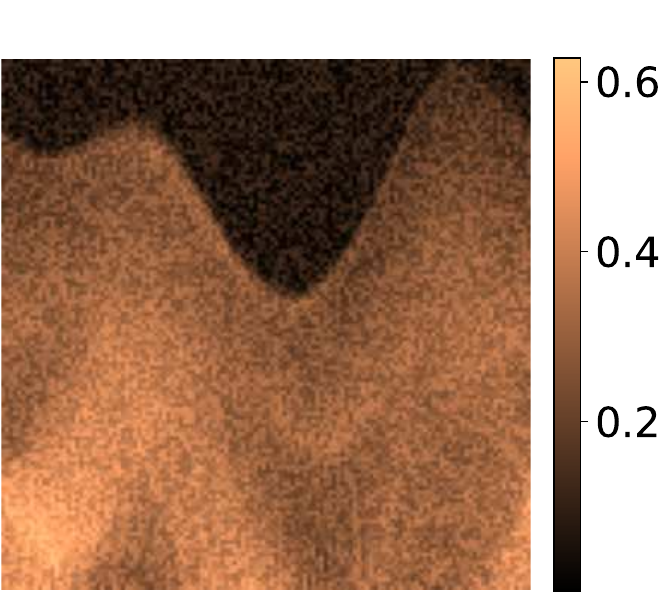} & \includegraphics[width=0.29\textwidth]{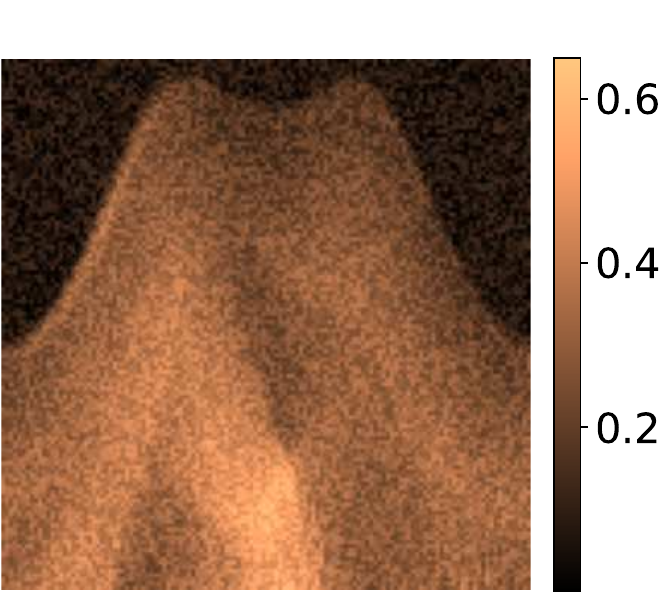}& \includegraphics[width=0.29\textwidth]{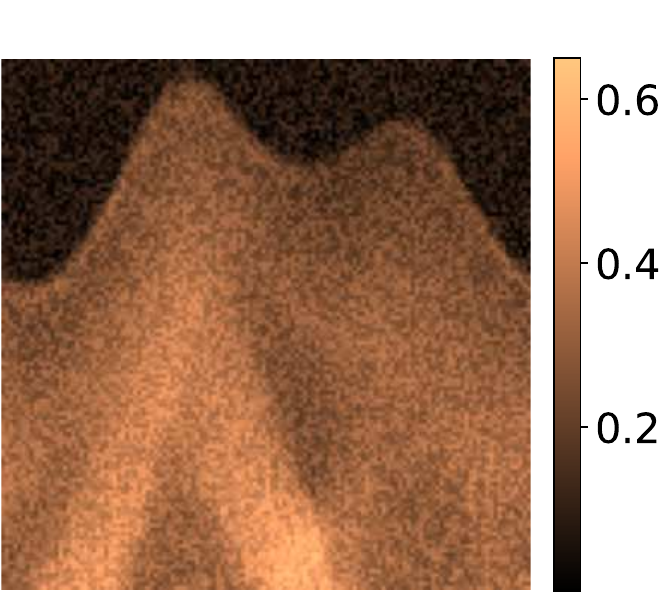} \tabularnewline
\rotatebox{90}{\textbf{UNet128c}} & \includegraphics[width=0.29\textwidth]{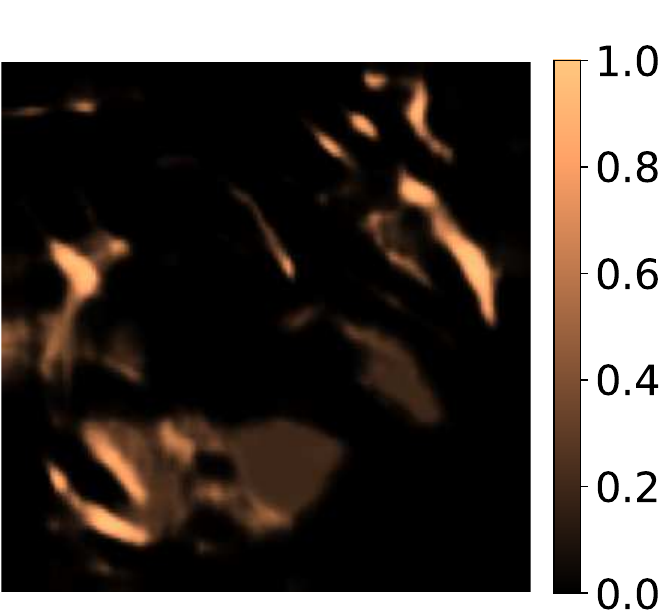} & \includegraphics[width=0.29\textwidth]{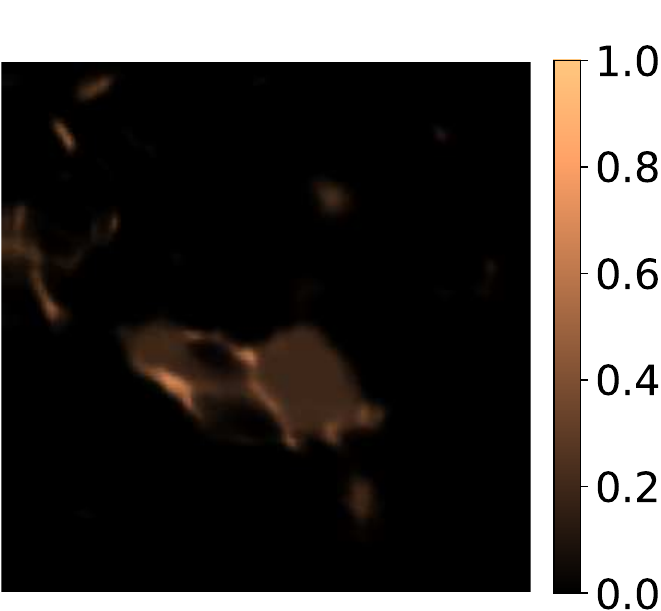}& \includegraphics[width=0.29\textwidth]{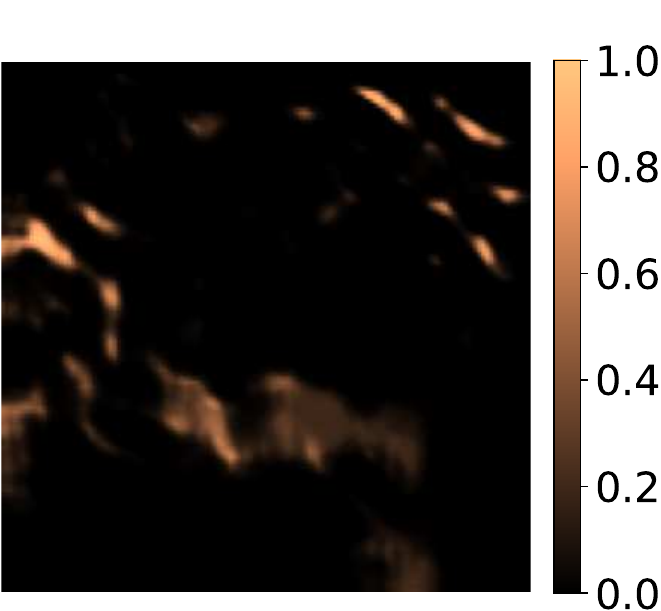} \tabularnewline
\rotatebox{90}{\textbf{TSVD}} & \includegraphics[width=0.27\textwidth]{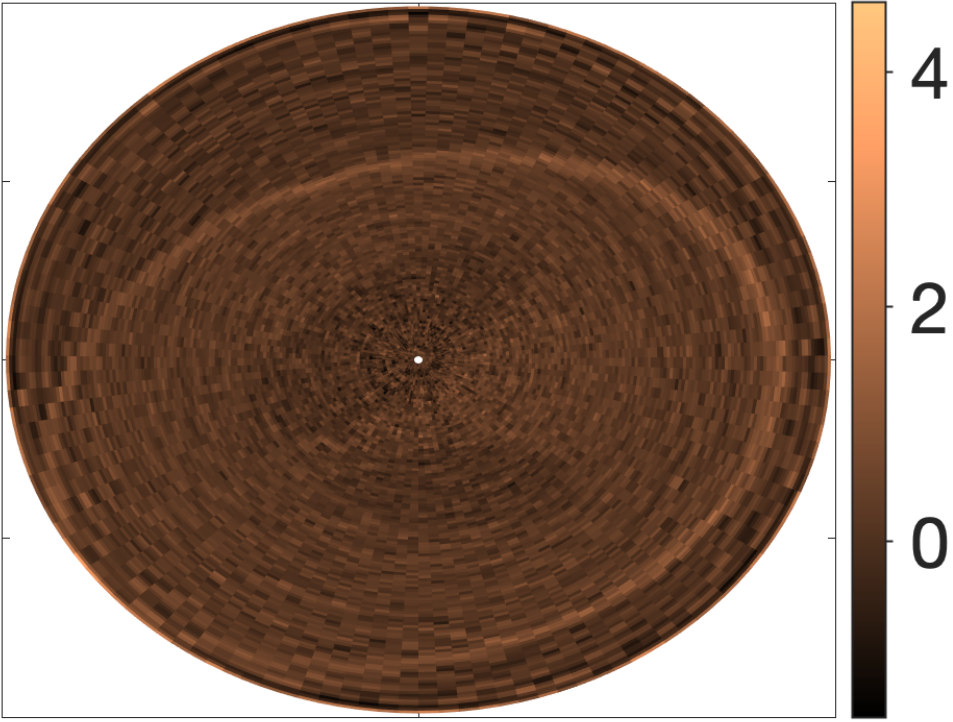} & \includegraphics[width=0.27\textwidth]{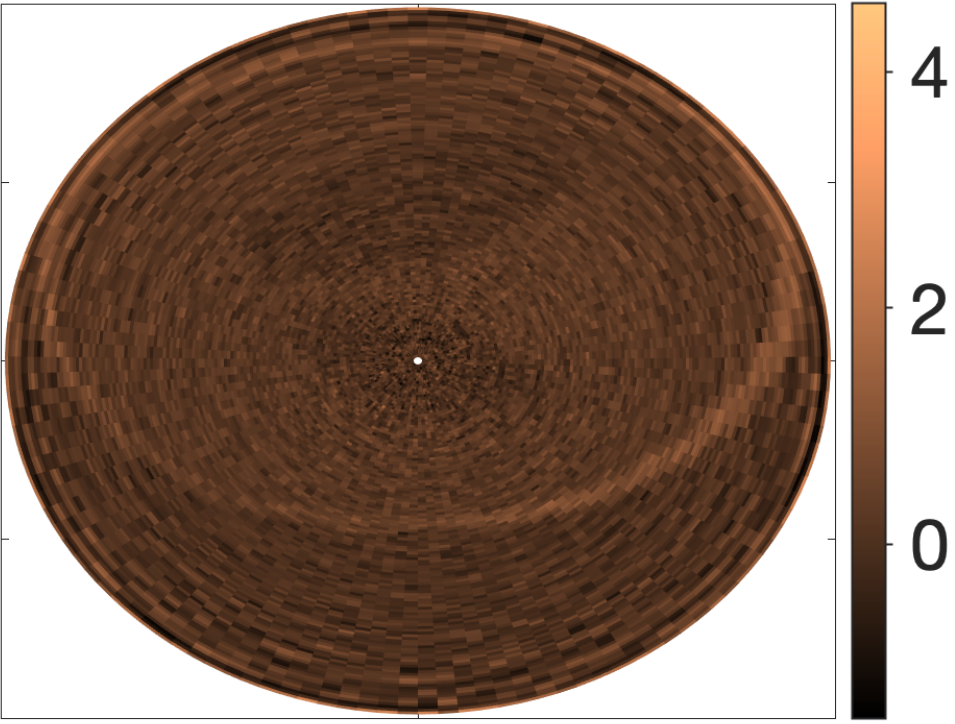}& \includegraphics[width=0.27\textwidth]{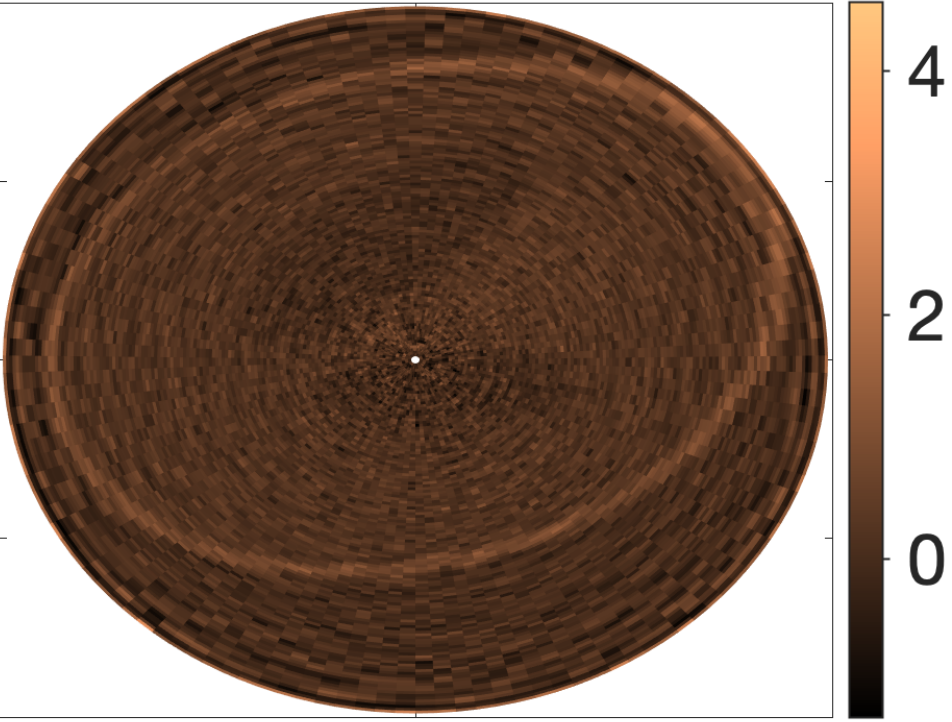}
\end{tabular}
\caption{Results for samples from test set Test128n15 with UNet128c and TSVD. }
\label{fig:NetA_QIII}
\end{figure}

However, UNet128c performs poorly when tested on Test128n5 containing samples with $5\%$ noise, as shown in Figure \ref{fig:NetA_QII}. The performance is even worse when the the network is tested with samples from set Test128n15, which has a higher measurement noise of $15\%$ (see Figure \ref{fig:NetA_QIII}). Although the more traditional TSVD is able to capture the underlying phantom structure with both levels of noise, the artefacting exacerbates with increasing levels of measurement noise. In fact, for $15\%$ measurement noise, the artefacting can lead to up to a $400\%$ error in reconstructing the phantom.      

\begin{figure}[!htbp]
\centering
\subfloat[Sample 1 from Test128n5]{\includegraphics[width=0.29\textwidth]{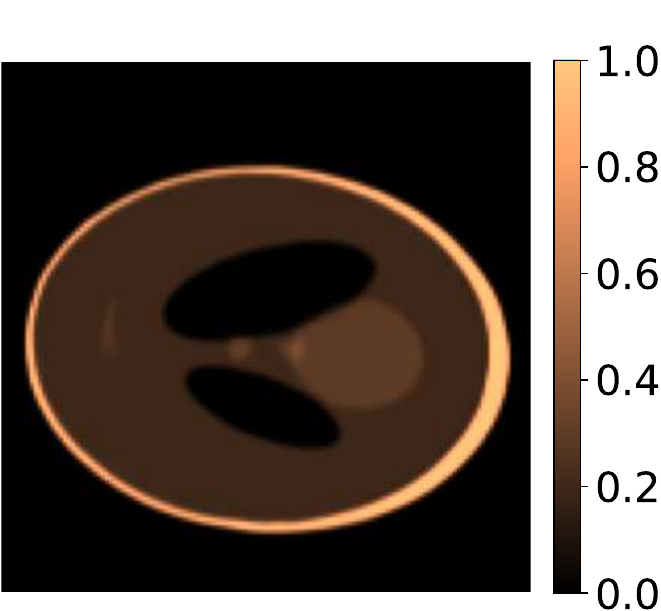}}\hfill 
\subfloat[Sample 2 from Test128n5]{\includegraphics[width=0.29\textwidth]{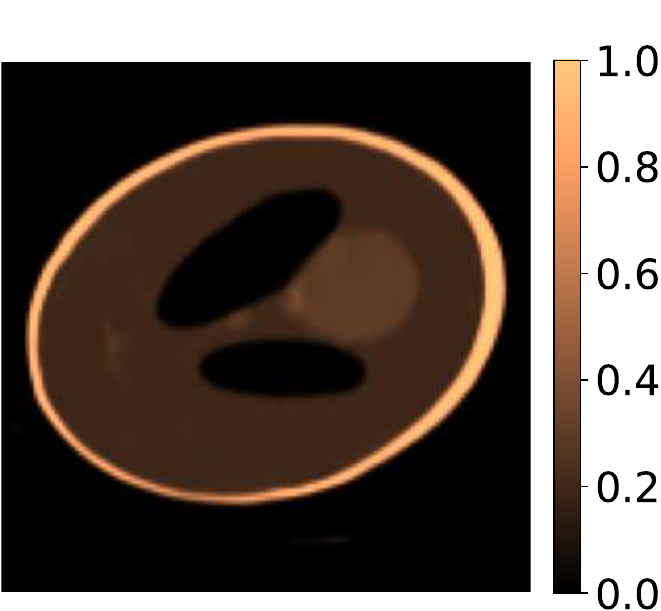}} \hfill
\subfloat[Sample 3 from Test128n5]{\includegraphics[width=0.29\textwidth]{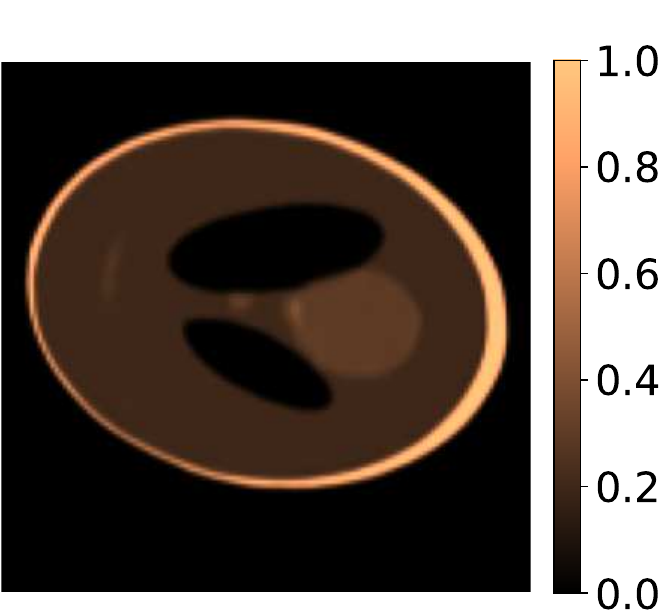}}\\
\subfloat[Sample 1 from Test128n15]{\includegraphics[width=0.29\textwidth]{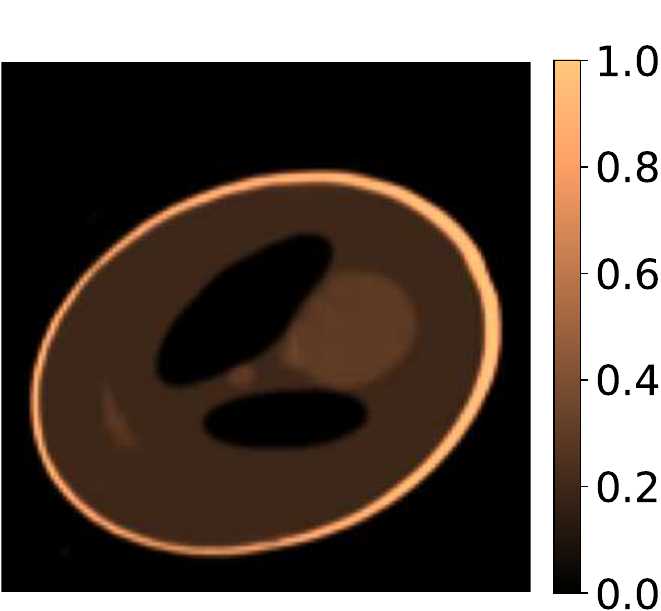}}\hfill 
\subfloat[Sample 2 from Test128n15]{\includegraphics[width=0.29\textwidth]{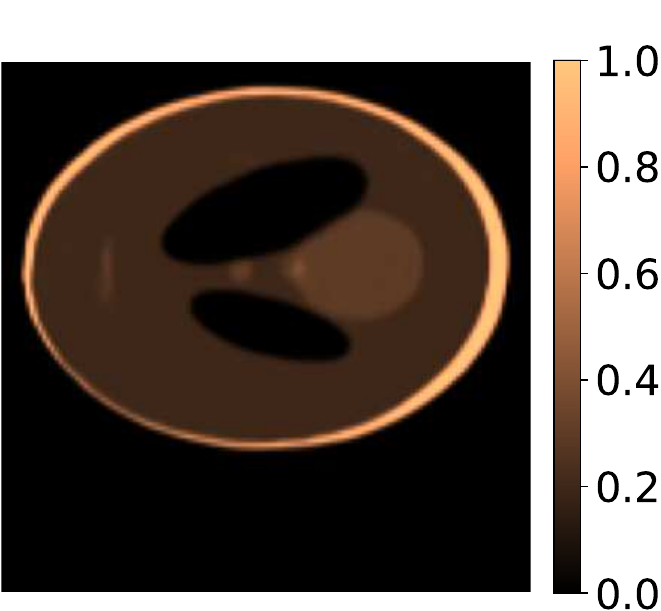}} \hfill
\subfloat[Sample 3 from Test128n15]{\includegraphics[width=0.29\textwidth]{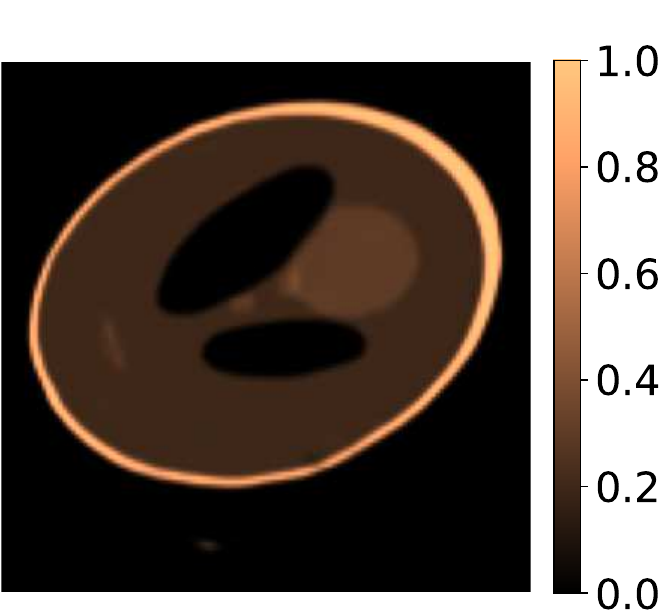}}\\
\caption{Reconstructed phantoms with UNet128cn15 for samples from test sets Test128n5 and Test128n15. Compare these with the true phantoms shown in Figures \ref{fig:NetA_QII} and \ref{fig:NetA_QIII}. }
\label{fig:NetB_QII_III}
\end{figure}

\begin{figure}[!htbp]
\centering
\subfloat[PSNR: Test128c]{\includegraphics[width=0.3\textwidth, height=0.3\textwidth]{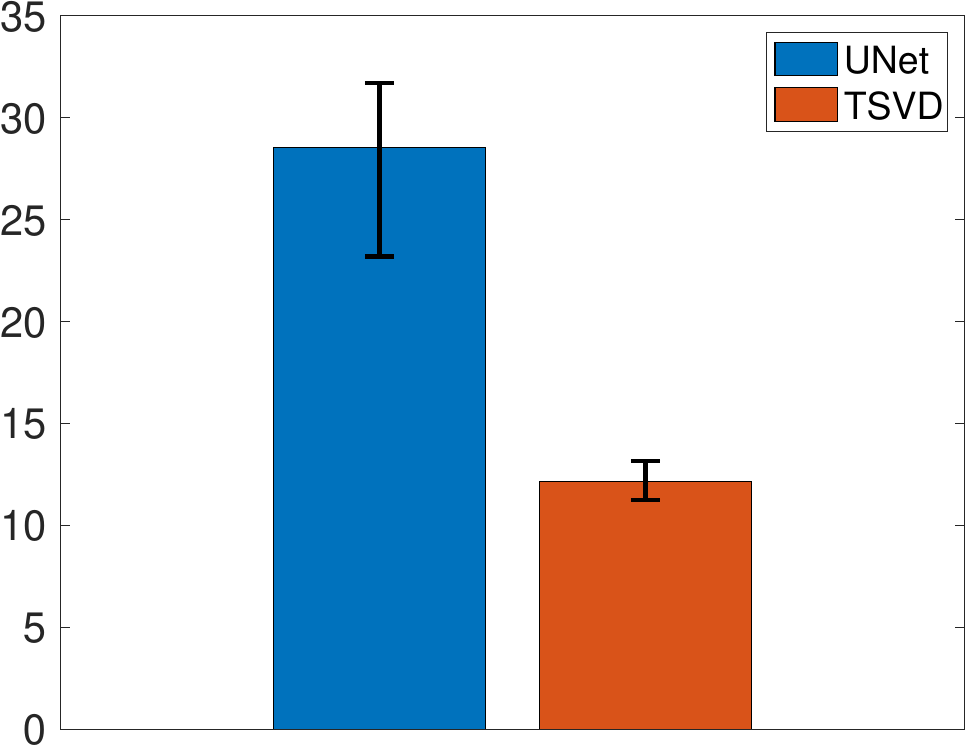}}\hfill
\subfloat[PSNR: Test128n5]{\includegraphics[width=0.3\textwidth, height=0.3\textwidth]{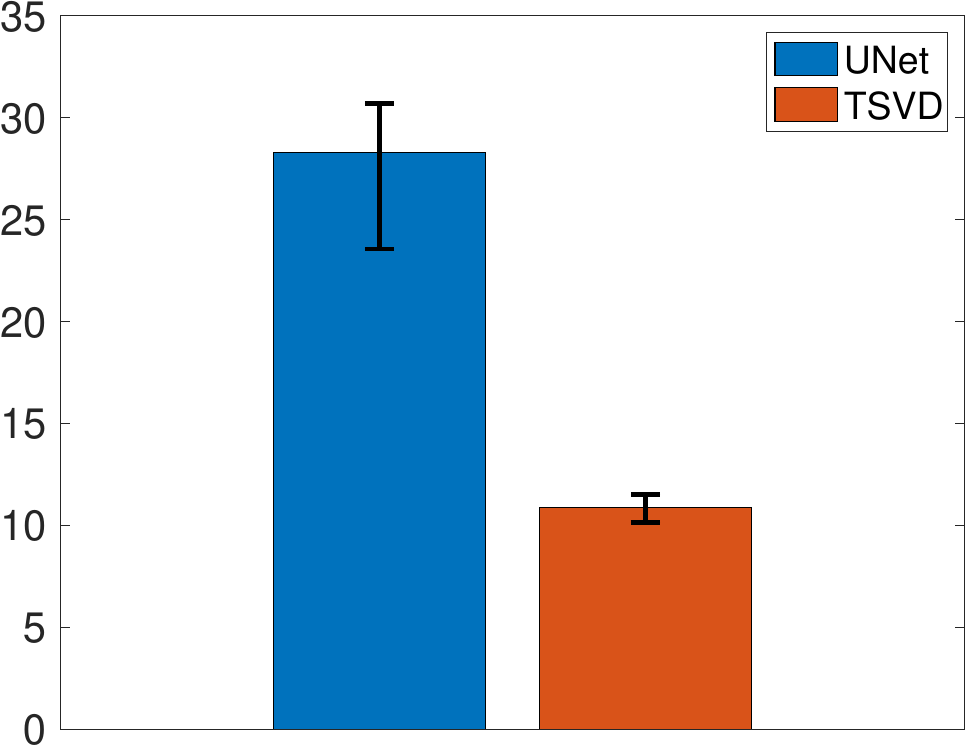}} \hfill
\subfloat[PSNR: Test128n15]{\includegraphics[width=0.3\textwidth, height=0.3\textwidth]{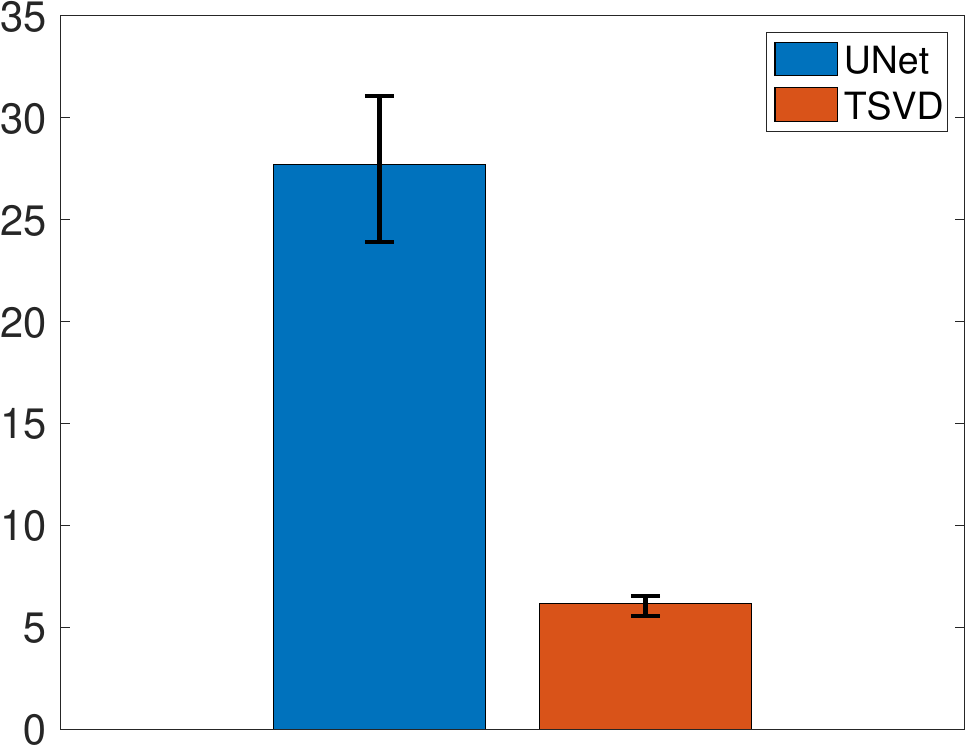}}\\
\subfloat[SSIM: Test128c]{\includegraphics[width=0.3\textwidth, height=0.3\textwidth]{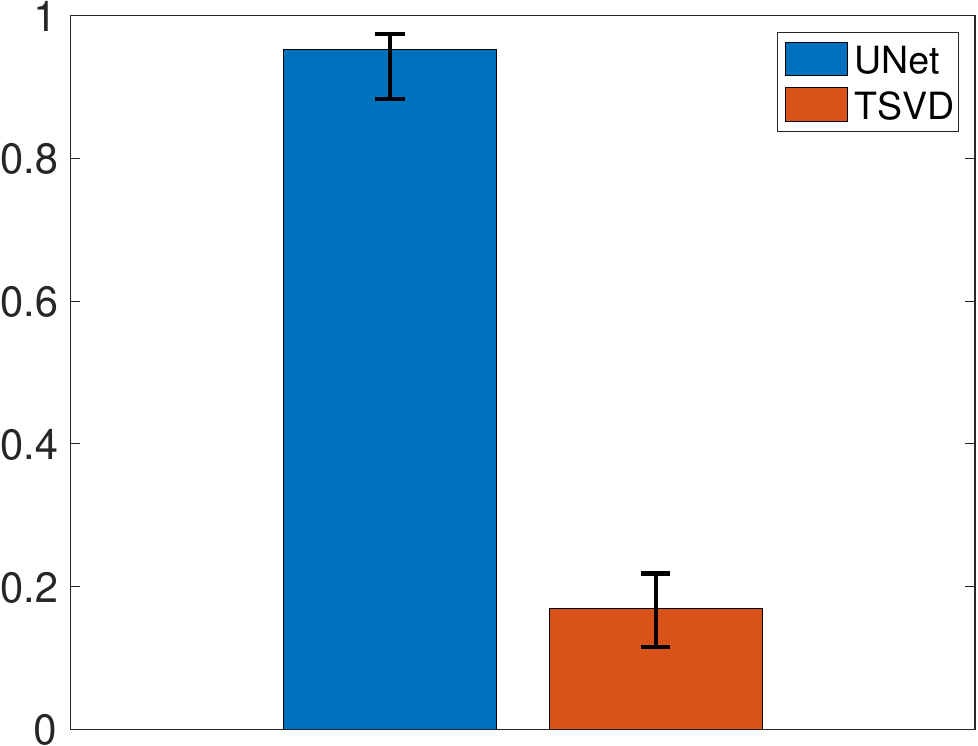}}\hfill
\subfloat[SSIM: Test128n5]{\includegraphics[width=0.3\textwidth, height=0.3\textwidth]{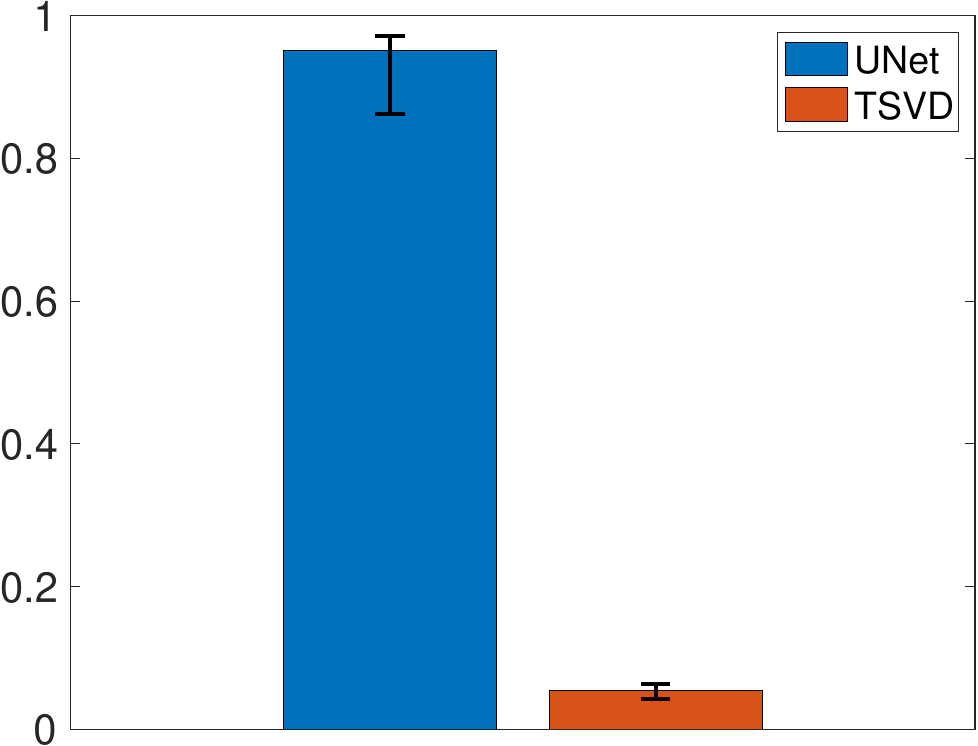}} \hfill
\subfloat[SSIM: Test128n15]{\includegraphics[width=0.3\textwidth, height=0.3\textwidth]{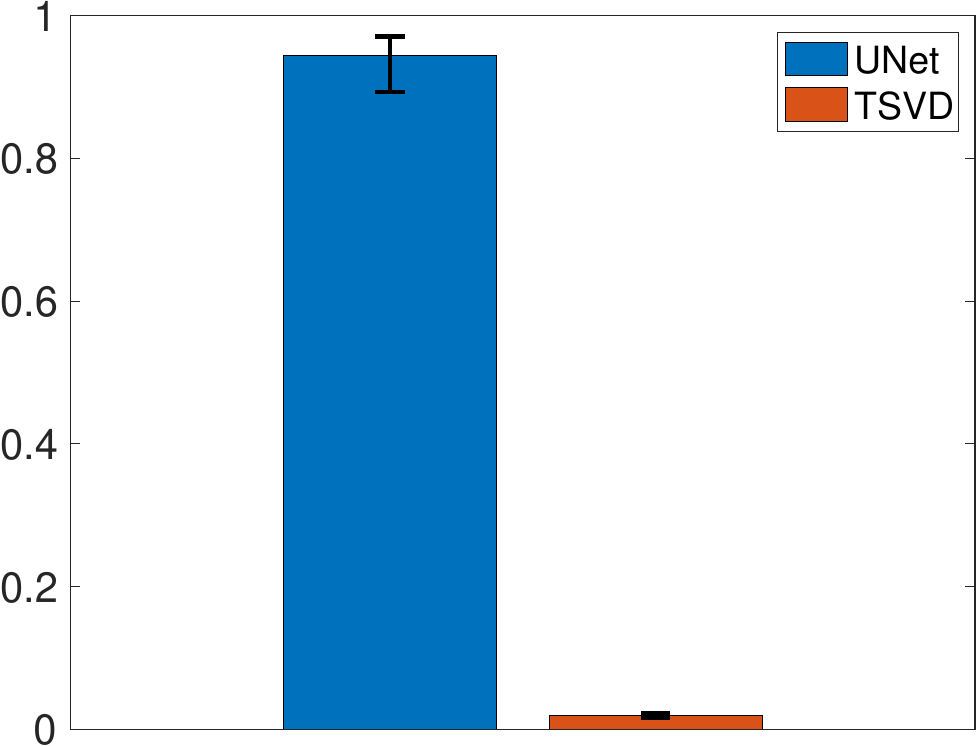}}\\
\caption{Mean PSNR and SSIM metric for TSVD and UNet128cn15. The mean is taken over all samples in the listed test set, while the horizontal bars denote $\pm$ standard deviation.}
\label{fig:metrics_unetB}
\end{figure}

In order to overcome the challenges faced by UNet128c, we train another U-Net on Train128cn15 which contains both both clean and noisy measurements (15\% noise). The training and validation losses for this new network, termed as UNet128cn15, are shown in Figure \ref{fig:losses}(b). Compared to UNet128c, we note that training with noisy training samples leads to better performance on the validation samples, i.e., the generalization is better. As shown in Figure \ref{fig:NetB_QII_III}, UNet128cn15 leads to an accurate and artefact-free recovery of the underlying phantom for samples from Test128n5 with a moderate $5\%$ noise, as well as for samples from Test128n15 with a higher $15\%$ noise. Note that the training set for UNet128cn15 did not contain samples with $5\%$ noise.
To further highlight the superiority of UNet128cn15 over the more traditional TSVD algorithm, we evaluate the mean peak signal-to-noise ratio (PSNR) and structural similarity index (SSIM) metrics, which are commonly used in computer vision applications to assess the quality of reconstructed/de-noised images. It is desirable to have a high value of PSNR, which is inversely related to the mean square error, and an SSIM close to unity. As shown in Figure \ref{fig:metrics_unetB}, both metrics are significantly higher for UNet128cn15 compared to TSVD. Moreover, while the metric values decrease for TSVD as more noise is added (increasing from 0\% to 15\%), the metrics remain fairly stable across all datasets for UNet128cn15. This demonstrates the robustness of the trained network.  

\subsection{Results for limited view measurements}
Next, we consider the limited view setup. We recall that for our experiments, this translates to a limited range of $[0,\pi]$ in the angular coordinate $\phi$ and a coarser sampling in the radial component $\rho$. The measurement tensor $\Y$ has a size of $64\times64$ , as opposed to $128 \times 128$ obtained with a full view measurement. Since $\Y$ is fed as input to the network, we cannot reuse the network trained for the full view setup. Further, motivated by the quality of reconstruction obtained with UNet128cn15 compared to UNet128c, we train the network using a dataset containing both noise-free samples as well as samples with maximal expected noise level (15\% in this example) to ensure robust performance of the network for the entire noise spectrum. Thus, the new U-Net is trained on Train64cn15, and termed as UNet64cn15. The training and validation losses are shown in Figure \ref{fig:losses}(c). Note that the both the loss curves take larger values as compared to the previous two networks. This is not unexpected as we are trying to reconstruct the phantom at the same resolution but with sparser measurement. However, we demonstrate below (qualitatively and quantitatively) that the reconstruction is still very accurate.

\begin{figure}[!htbp]
\centering
\begin{tabular}{ m{0.2cm} >{\centering}m{3.2cm} >{\centering}m{3.2cm} >{\centering}m{3.2cm} }
& \textbf{Test Sample 1} & \textbf{Test Sample 2} & \textbf{Test Sample 3} \tabularnewline
\rotatebox{90}{\textbf{True $\X$}} & \includegraphics[width=0.29\textwidth]{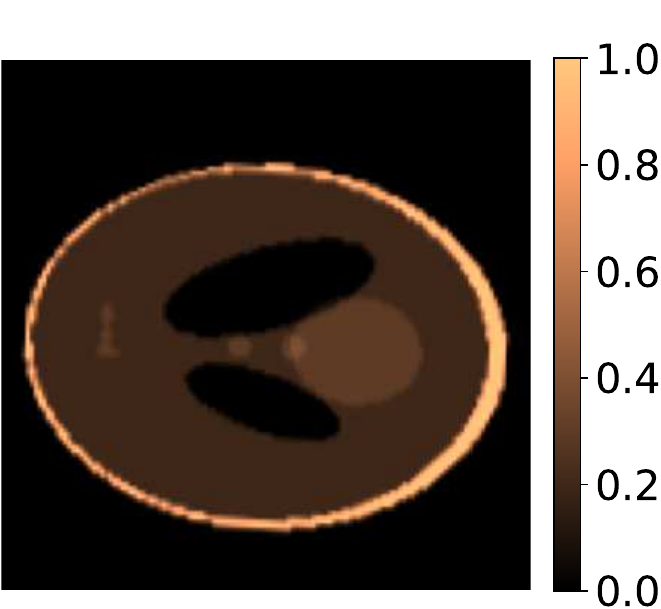} & \includegraphics[width=0.29\textwidth]{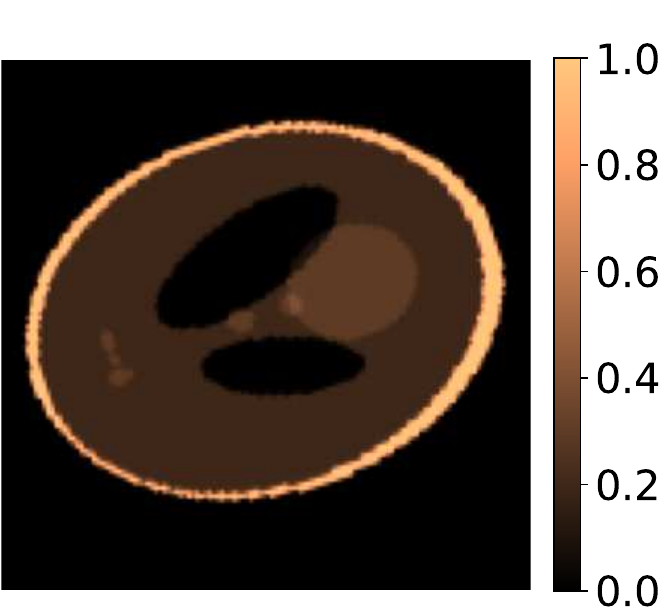}& \includegraphics[width=0.29\textwidth]{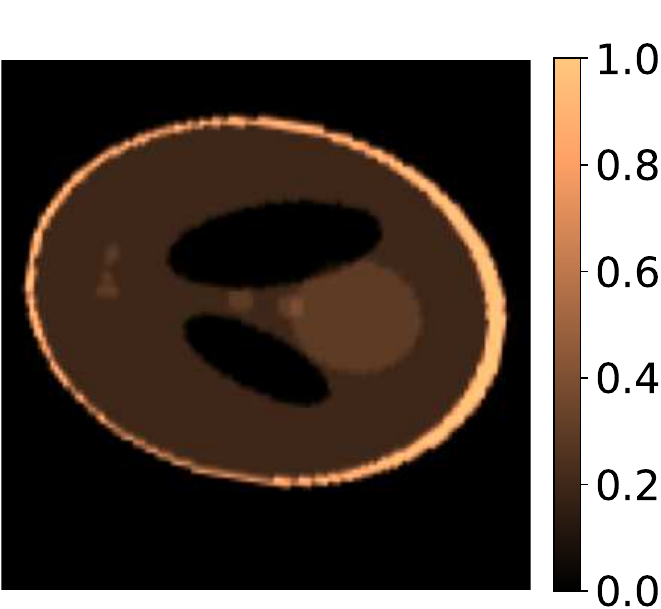} \tabularnewline
\rotatebox{90}{\textbf{Measurement}} & \includegraphics[width=0.29\textwidth]{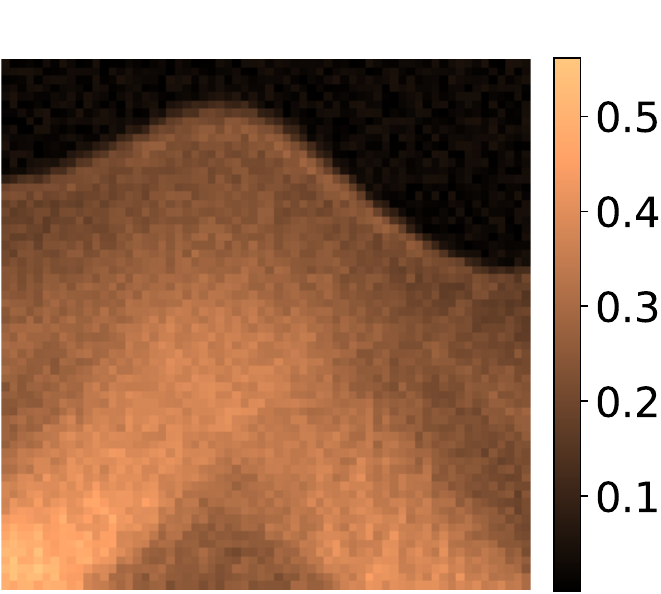} & \includegraphics[width=0.29\textwidth]{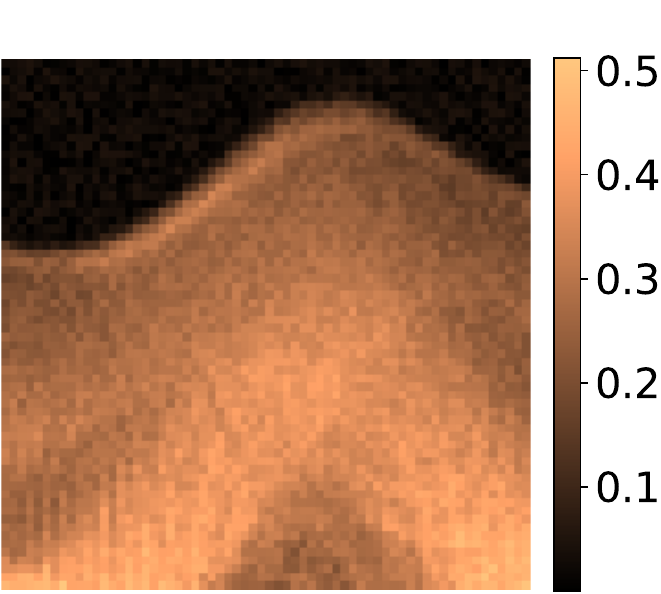}& \includegraphics[width=0.29\textwidth]{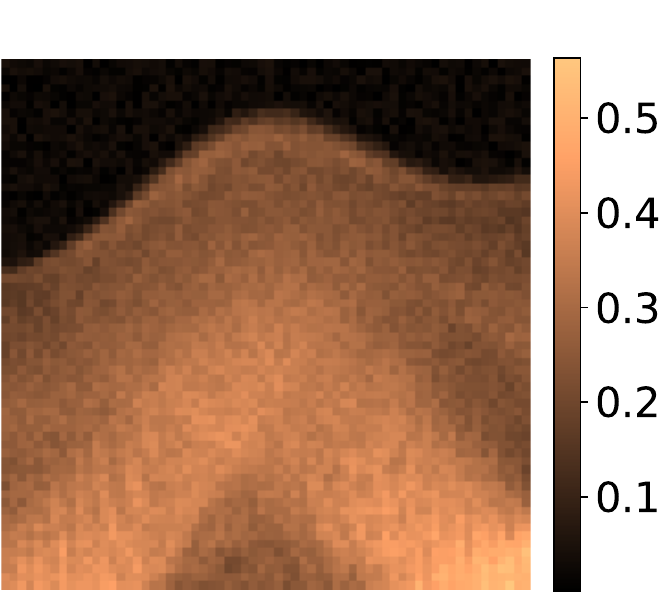} \tabularnewline
\rotatebox{90}{\textbf{UNet64cn15}} & \includegraphics[width=0.29\textwidth]{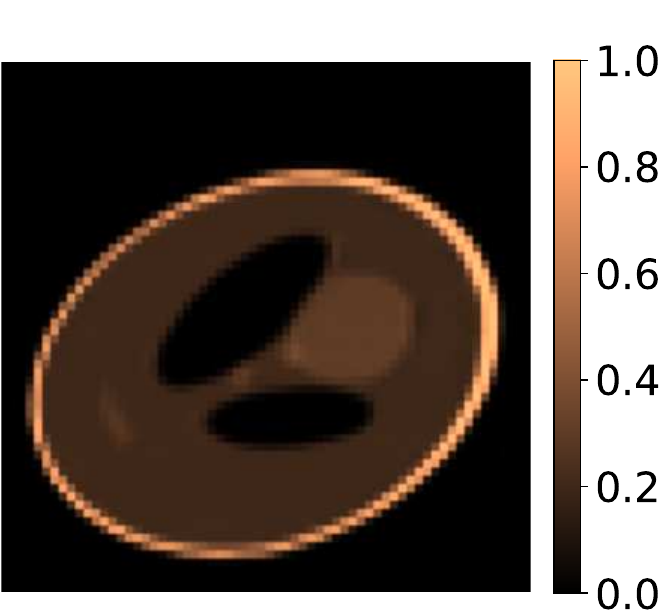} & \includegraphics[width=0.29\textwidth]{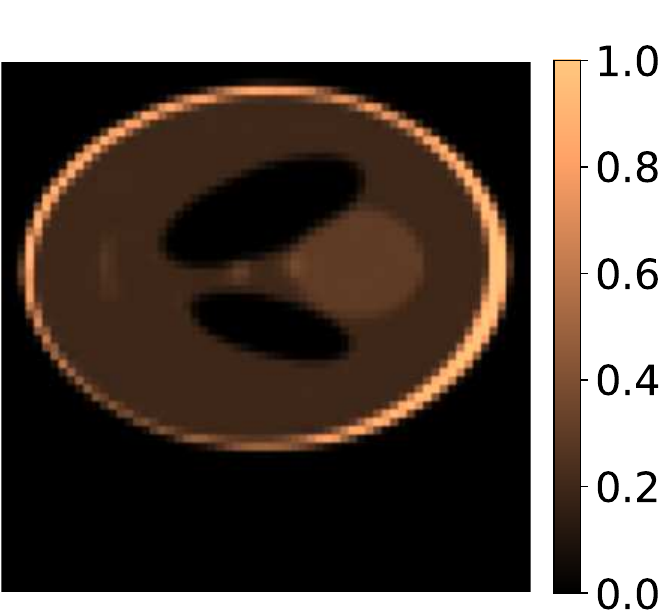}& \includegraphics[width=0.29\textwidth]{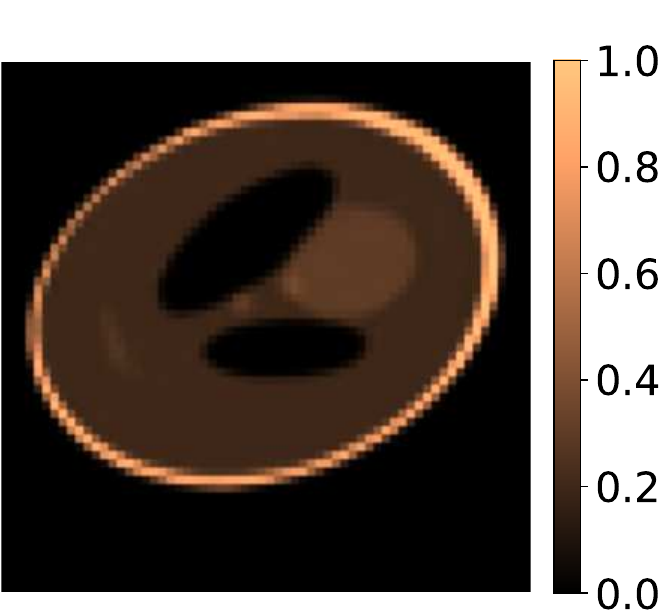} \tabularnewline
\rotatebox{90}{\textbf{TSVD}} & \includegraphics[width=0.28\textwidth]{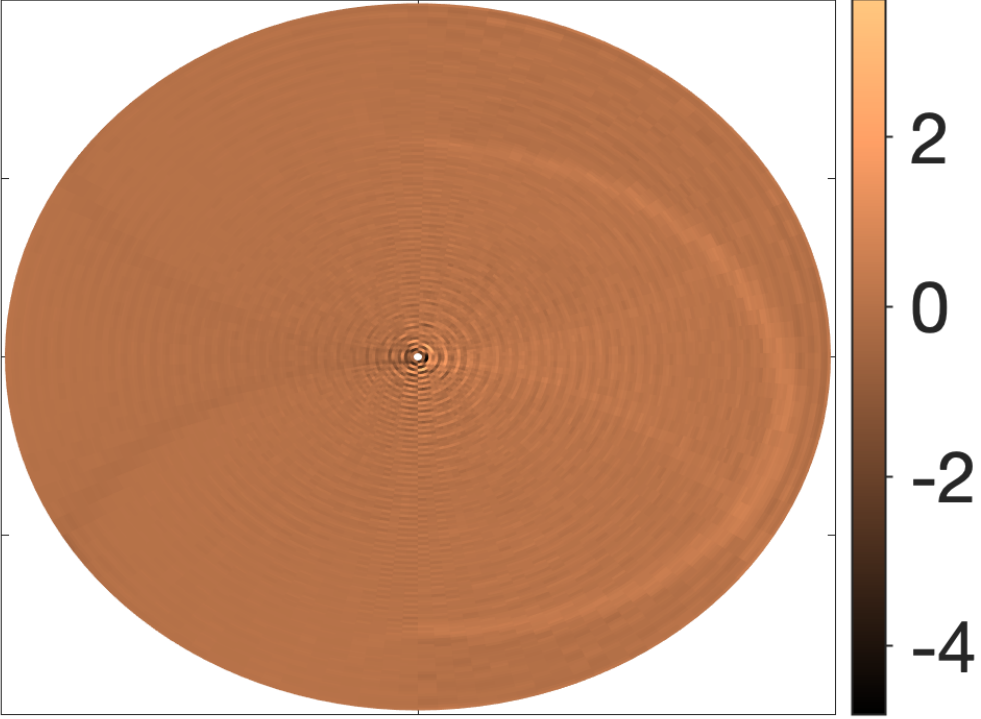} & \includegraphics[width=0.28\textwidth]{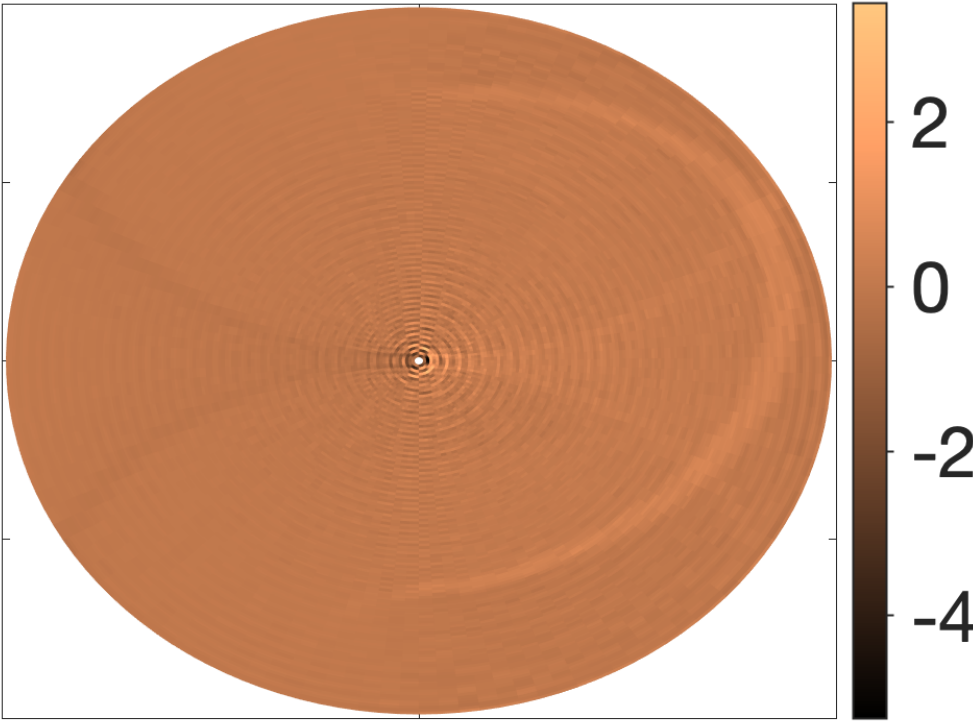}& \includegraphics[width=0.28\textwidth]{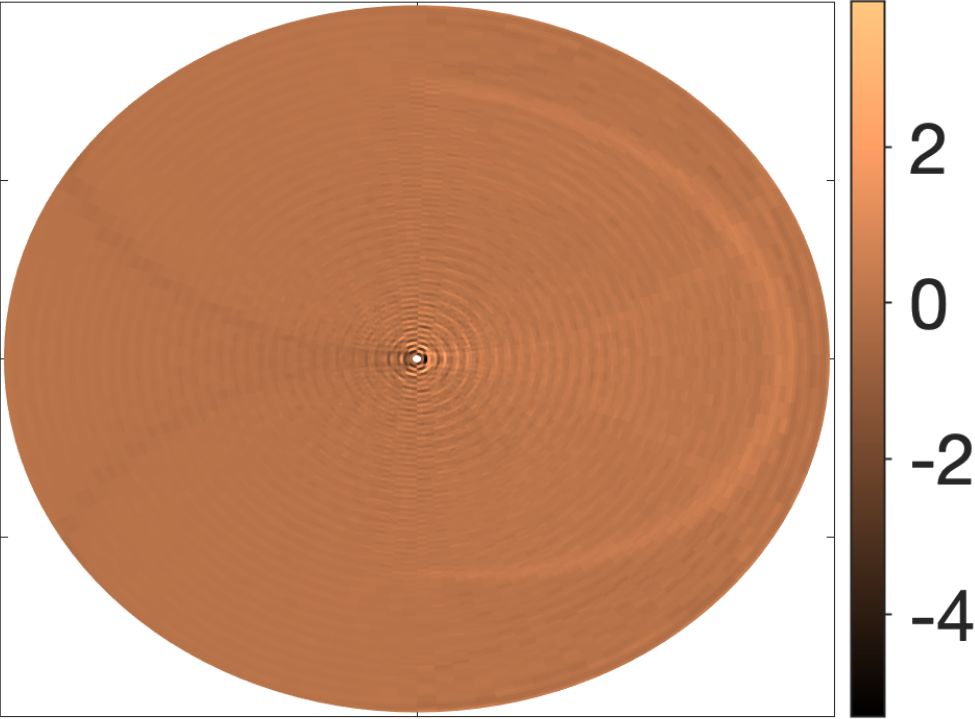}
\end{tabular}
\caption{Results for samples from test set Test64n5 with UNet64cn15 and TSVD. }

\label{fig:NetC_QIV}
\end{figure}

\begin{figure}[!htbp]
\centering
\begin{tabular}{ m{0.2cm} >{\centering}m{3.2cm} >{\centering}m{3.2cm} >{\centering}m{3.2cm} }
& \textbf{Test Sample 1} & \textbf{Test Sample 2} & \textbf{Test Sample 3} \tabularnewline
\rotatebox{90}{\textbf{True $\X$}} & \includegraphics[width=0.29\textwidth]{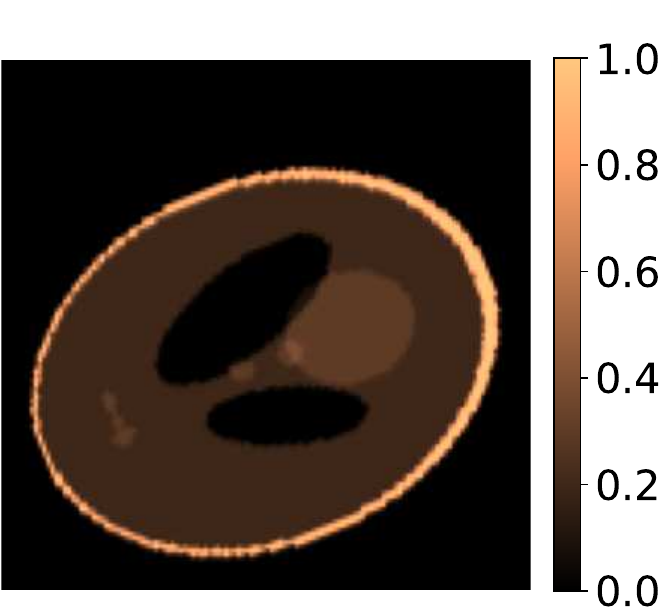} & \includegraphics[width=0.29\textwidth]{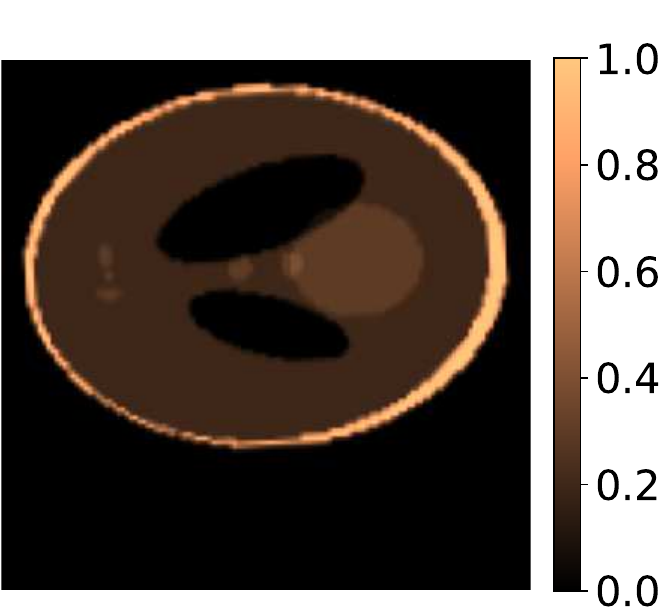}& \includegraphics[width=0.29\textwidth]{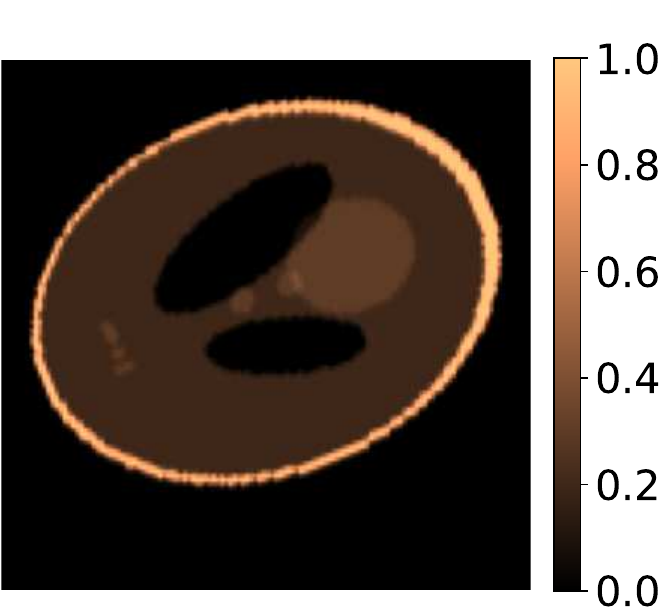} \tabularnewline
\rotatebox{90}{\textbf{Measurement}} & \includegraphics[width=0.29\textwidth]{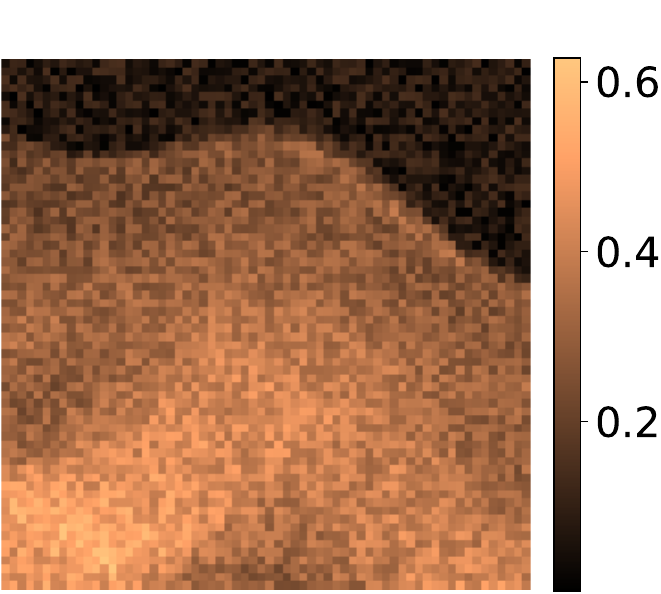} & \includegraphics[width=0.29\textwidth]{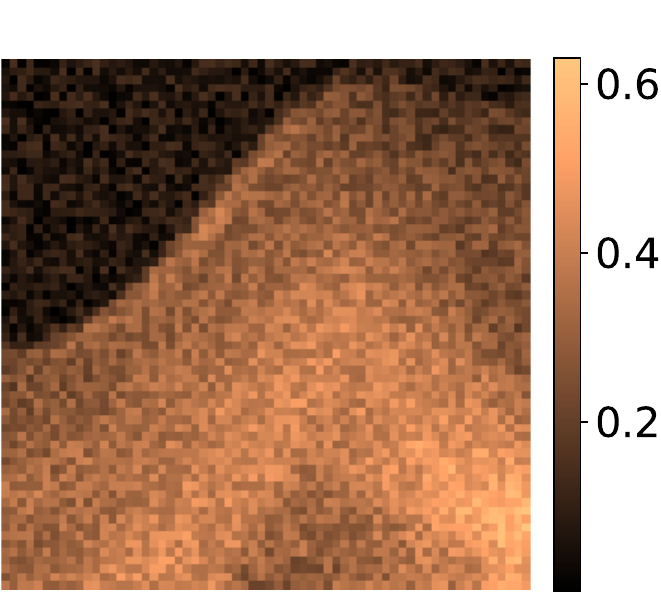}& \includegraphics[width=0.29\textwidth]{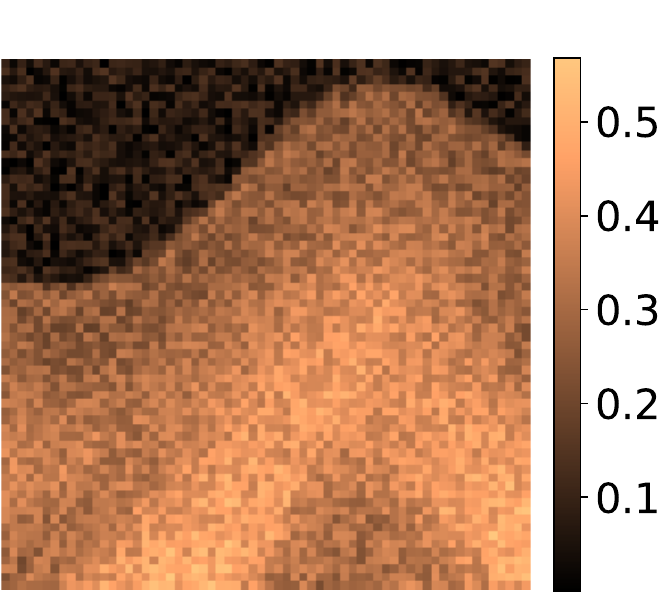} \tabularnewline
\rotatebox{90}{\textbf{UNet64cn15}} & \includegraphics[width=0.29\textwidth]{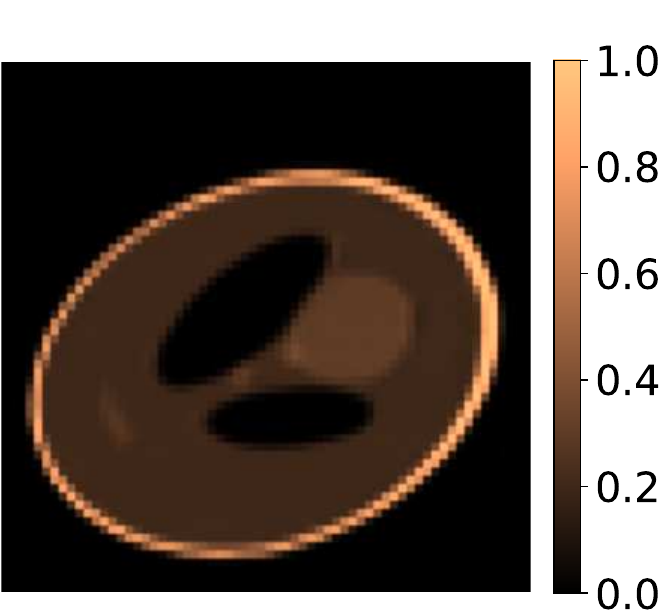} & \includegraphics[width=0.29\textwidth]{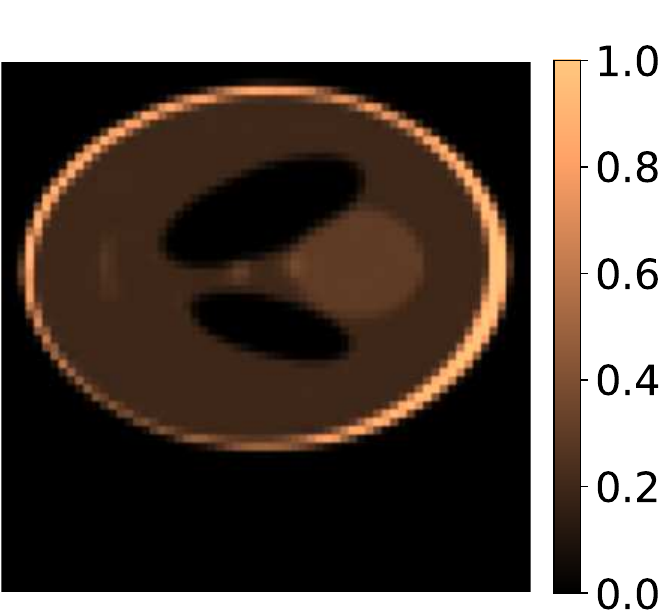}& \includegraphics[width=0.29\textwidth]{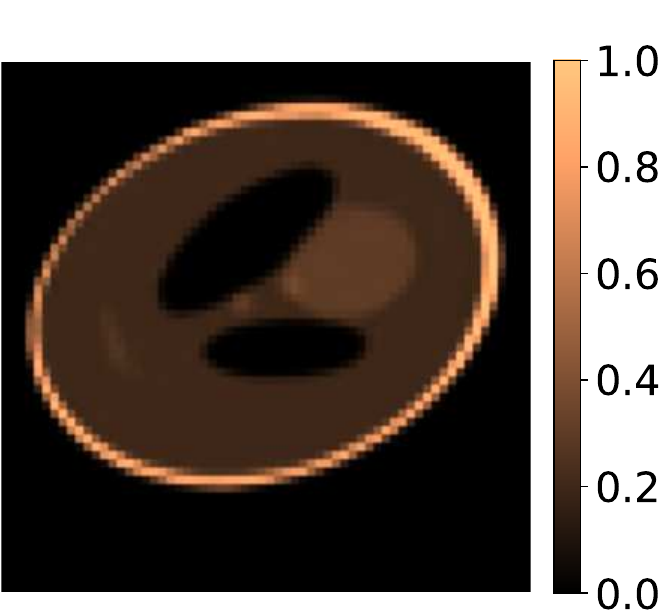} \tabularnewline
\rotatebox{90}{\textbf{TSVD}} & \includegraphics[width=0.28\textwidth]{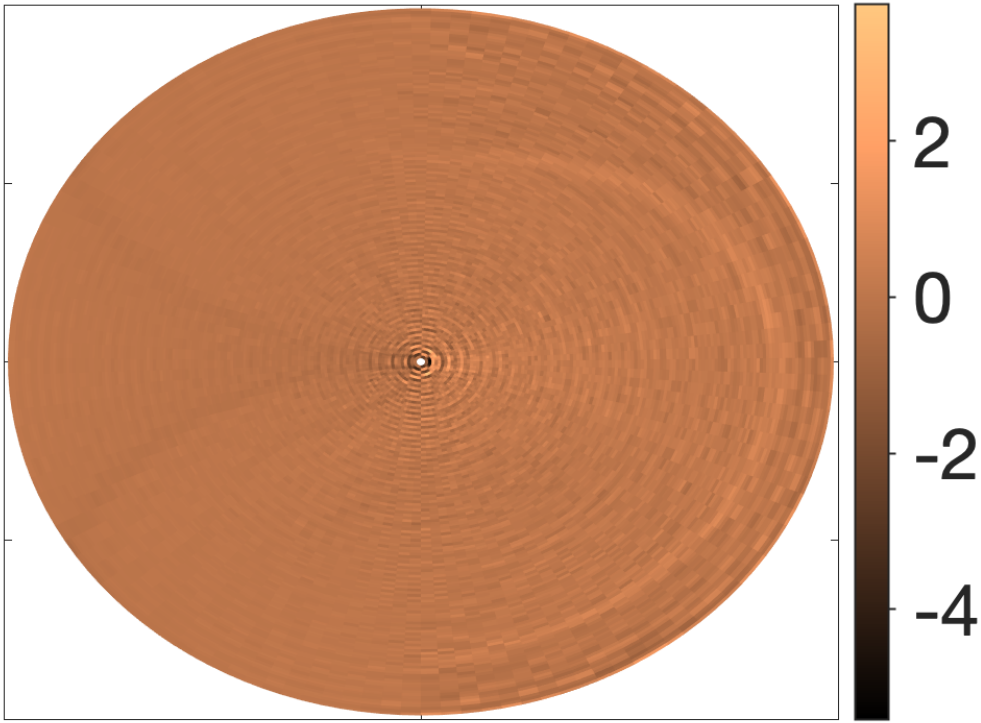} & \includegraphics[width=0.28\textwidth]{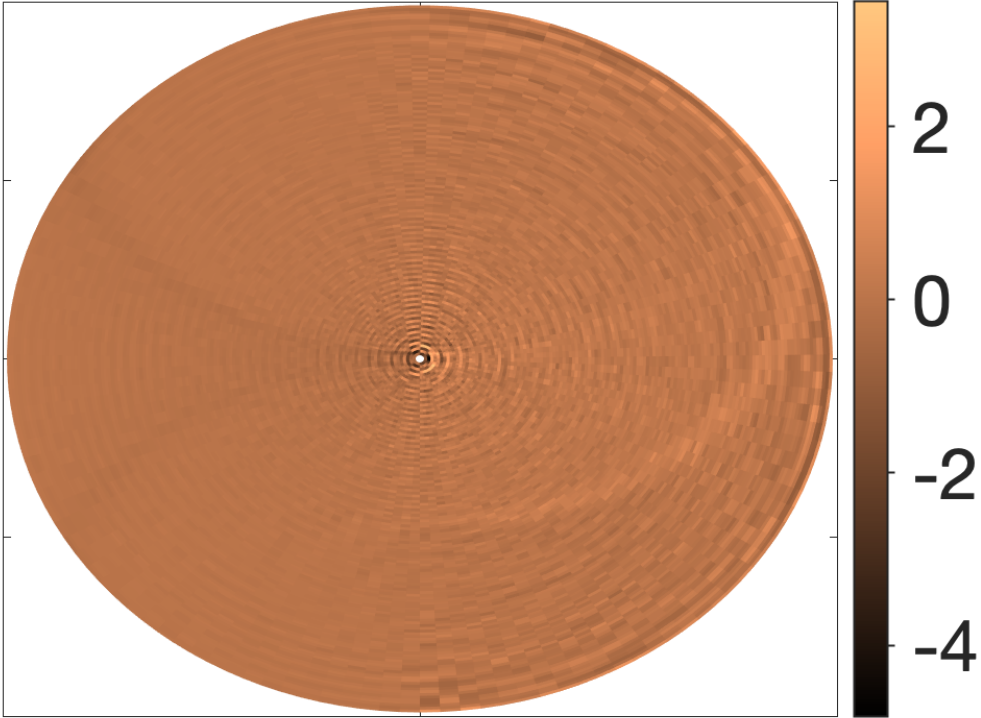}& \includegraphics[width=0.28\textwidth]{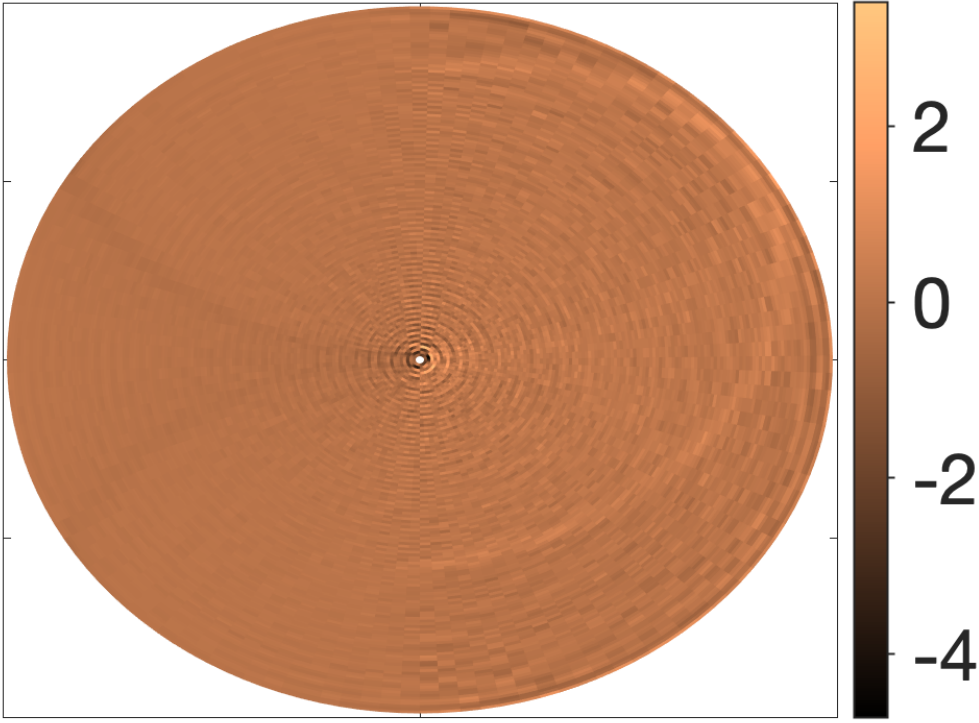}
\end{tabular}
\caption{Results for samples from test set Test64n15 with UNet64cn15 and TSVD. }
\label{fig:NetC_QV}
\end{figure}

\begin{figure}[!htbp]
\centering
\subfloat[PSNR: Test64n5]{\includegraphics[width=0.3\textwidth, height=0.3\textwidth]{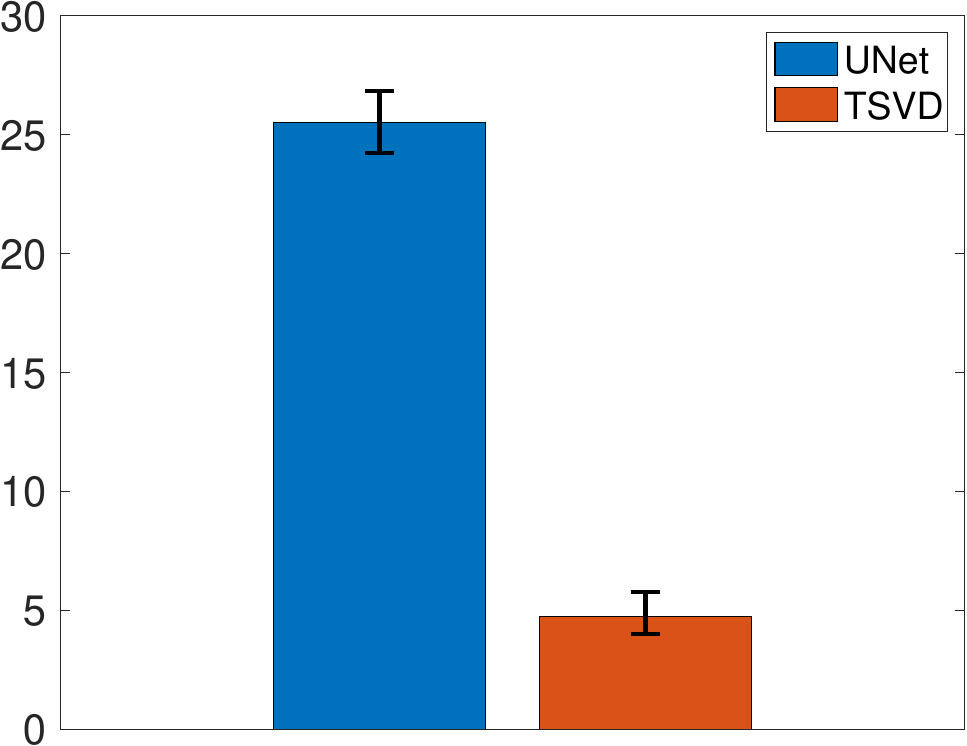}} \hspace{0.5cm}
\subfloat[PSNR: Test64n15]{\includegraphics[width=0.3\textwidth, height=0.3\textwidth]{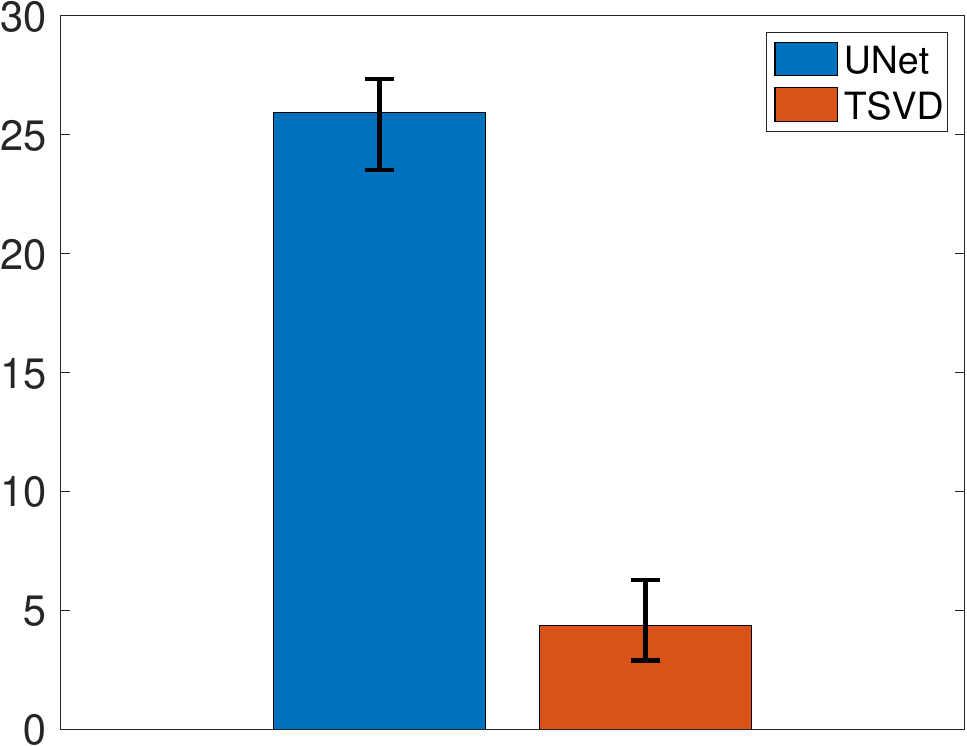}}\\
\subfloat[SSIM: Test64n5]{\includegraphics[width=0.3\textwidth, height=0.3\textwidth]{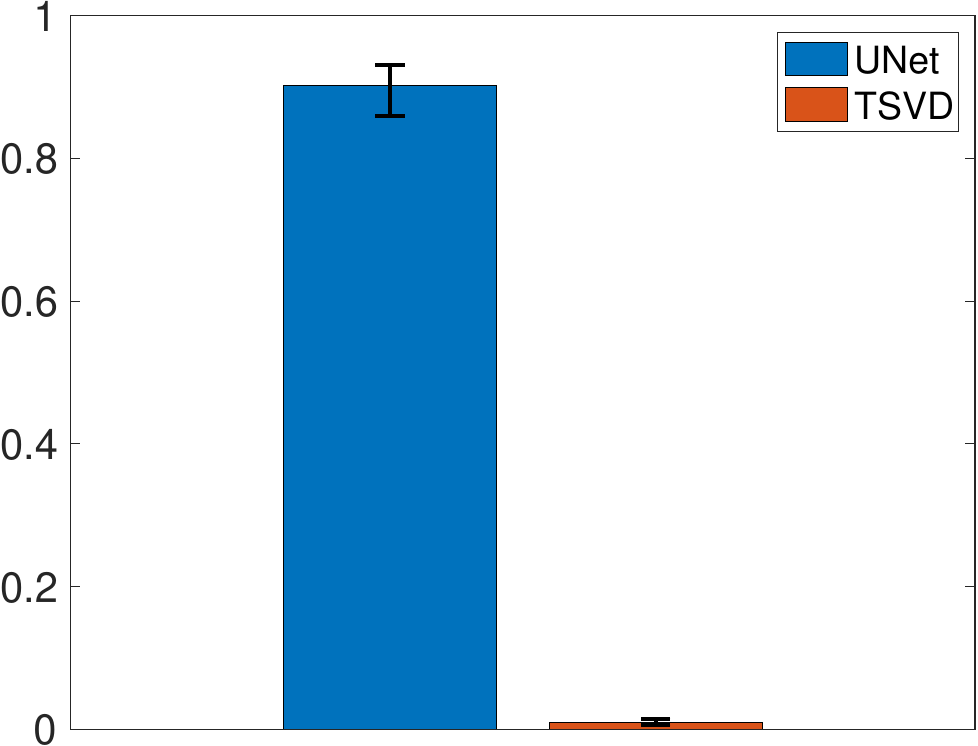}} \hspace{0.5cm}
\subfloat[SSIM: Test64n15]{\includegraphics[width=0.3\textwidth, height=0.3\textwidth]{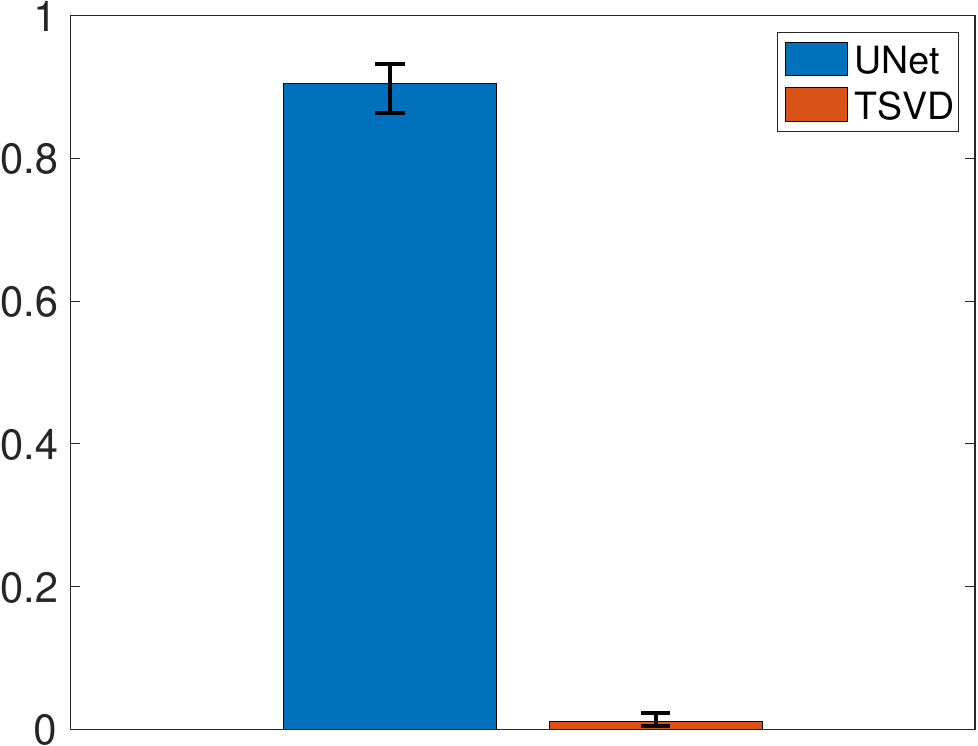}}
\caption{Mean PSNR and SSIM metric for TSVD and UNet64cn15. The mean is taken over all samples in the listed test set, while the horizontal bars denote $\pm$ standard deviation.}
\label{fig:metrics_unetC}
\end{figure}

We test UNet64cn15 on Test64n5 containing moderate noise of 5\% and Test64n15 containing maximal noise of 15\%, with the predictions show in Figures \ref{fig:NetC_QIV} and \ref{fig:NetC_QV}, respectively. The reconstructions are accurate and artefact-free, as opposed to the those obtained with TSVD which are severely corrupted by artefacts to the point that it is hard to discern the underlying phantom structure. The mean PSNR and SSIM metric shown in Figure \ref{fig:metrics_unetC} re-enforce the superior and robust performance of the network-based inversion as compared to the traditional TSVD method.  

\section{Conclusions}\label{sec:conclusions}
In this paper, we presented a deep learning-based computational algorithm for inversion of circular Radon transforms in the partial radial setup, arising in PAT under the constant sound speed assumption. We first demonstrated that the only available traditional algorithm for this setup, the truncated singular value decomposition-based method, lead to severe artificating which rendered the reconstructed field as unusable. To overcome this computational bottleneck, we trained a ResBlock-based U-Net to recover the inferred field, which directly operated on the measured data. Numerical results with augmented Shepp-Logan phantoms in the presence of noisy full and limited data demonstrate the superiority of the proposed algorithm. Future work in this direction will be to generalize the proposed U-Net algorithm to various PAT setups with non-constant sound speed and different types of available acoustic measurements like flux measurements. Furthermore, based on the strategies proposed in \cite{Vu2020,adler2018deep,ray2022}, we will consider quantifying the uncertainty in the recovered field by posing the problem in the Bayesian setup. Such a reliability estimate is critical in situations where high-stake decisions need to made based on the reconstruction.

\section*{Acknowledgements} S. Roy was supported by the US National Science Foundation Grant No: DMS-2309491 and the University of Texas at Arlington, Research Enhancement Program Grant No: 2022-605.

\appendix

\section{U-Net blocks}\label{app:unet_blocks}
\begin{figure}[htbp]
\centering
\subfloat[Res. Block]{\includegraphics[width=0.8\textwidth]{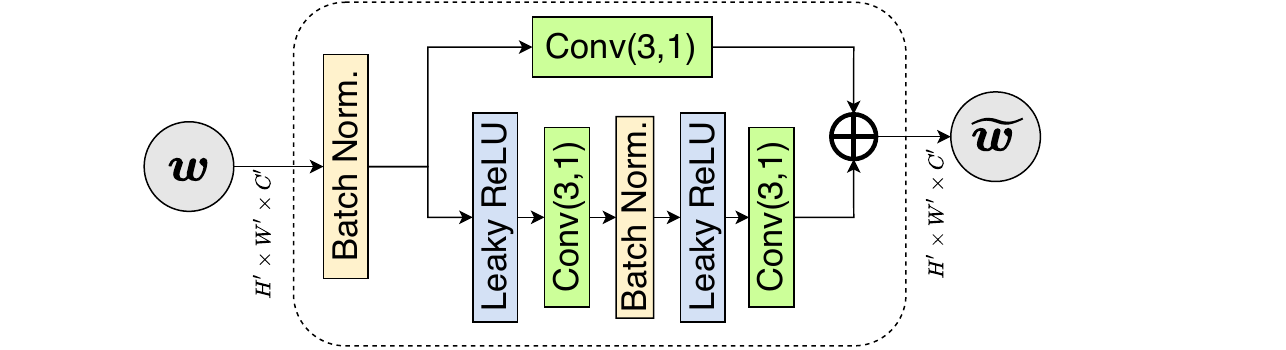}}\\
\subfloat[Down(p,k)]{\includegraphics[width=0.55\textwidth]{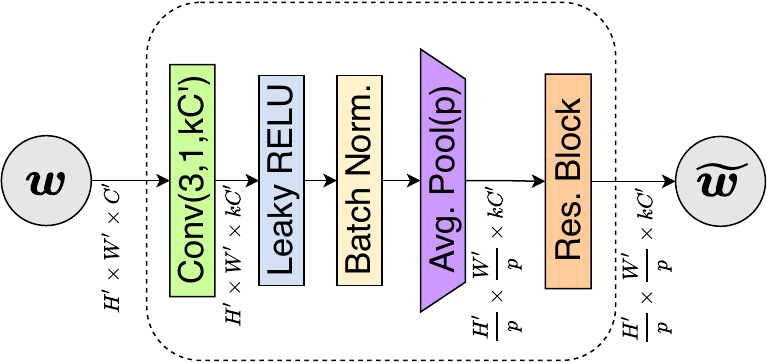}}\\
\subfloat[Up(p,k)]{\includegraphics[width=0.6\textwidth]{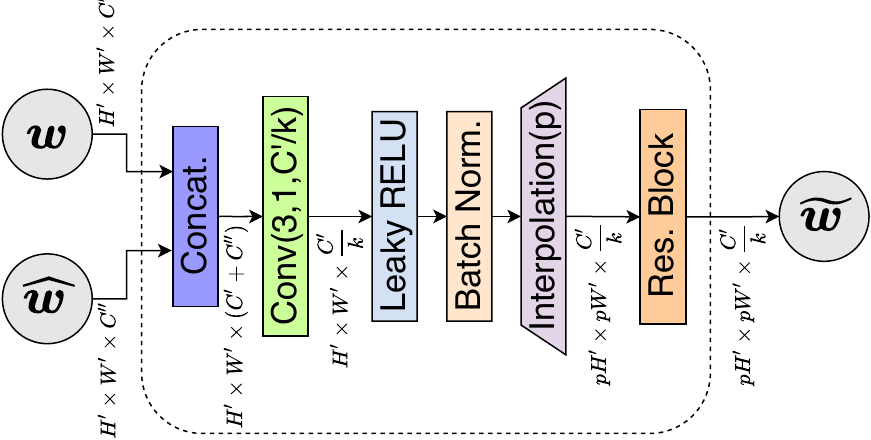}}
\caption{Key components used to build the U-Nets.}
\label{fig:unet_blocks}
\end{figure}

The U-Nets considered in the present work are based on 2D convolution layers and residual blocks. We describe the key components of the architecture below:

\begin{itemize}

    \item {\tt Conv($n,s,k$)} denotes a 2D convolution with $k$ filters of size $n$ and stride $s$. We apply a reflective padding of width 1 in the spatial dimensions of the input before applying the convolution, whenever $n>1$. If the third argument $k$ is absent, the number of filter is set to be equal to the number of channels in the input tensor.
    \item The Leaky ReLU with parameter $\alpha$ is as the non-linear activation function. The Sigmoid output function is used at the end of the U-Net, which ensures the pixel-wise values of the output are in the range $[0,1]$.
    \item Batch normalization is used to standardize the outputs from various layers and enhance the training \cite{batchnorm}.
    \item {\tt Res. Block} denotes a residual block, which is shown in Figure \ref{fig:unet_blocks}(a). It passes the input through two branches and adds the two outputs together. Note that the residual block preserves the shape of the input tensor.
    
    \item {\tt Down($p$,$k$)} denotes the down-sampling block shown in Figure \ref{fig:unet_blocks}(b). Here $p$ denotes the factor by which the input spatial resolution is reduced, which is achieved using a 2D average pooling denoted by {\tt Avg. Pool($p$)}. Further, $k$ denotes the factor by which the number of input channels increases.
    
    \item {\tt Up($p$,$k$)} denotes the up-sampling block shown in Figure \ref{fig:unet_blocks}(c).
    It receives the output $\boldsymbol{w}$ from the previous block and concatenates it (unless specified) with the output $\widehat{\boldsymbol{w}}$ of a down-sampling block of the same spatial size through a skip connection. The up-sampling block reduces the number of output channels to $C^\prime/k$, where $C^\prime$ denotes the number of channels in the input $\boldsymbol{w}$. The spatial resolution is increased by a factor of $p$ using 2D nearest neighbour interpolation, denoted by {\tt Interpolation(p)}. 
\end{itemize}    

\bibliographystyle{IEEEtran}
\bibliography{ref}


\end{document}